\documentclass[fleqn,leqno]{article}
\usepackage{fullname}
\usepackage{lingm}
\usepackage{avm}
\usepackage{graphicx}
\usepackage{pst-tree}
\usepackage{rotating}
\usepackage{palatino}
\usepackage{a4wide}
\newenvironment{numbered}{
  \begin{equation}\begin{minipage}[t]{.9\textwidth}}{%
  \end{minipage}\end{equation}}

\newenvironment{lnumbered}[1]{
  \begin{equation}\label{#1}\begin{minipage}[t]{.9\textwidth}}{%
  \end{minipage}\end{equation}}

\avmvskip{.1ex}
\avmhskip{.5em}
\avmoptions{center}
\newcommand{\ovis}{OVIS}
\newcommand{\ovistwo}{OVIS2}
\title{Robust Grammatical Analysis for Spoken Dialogue Systems}
\author{Gertjan van Noord, Gosse Bouma, Rob Koeling, Mark-Jan Nederhof\\
        Alfa-Informatica \& BCN\\
        University of Groningen}

\date{June 1998\\
      Accepted for Journal of Natural Language Engineering}

\begin{document}

\label{firstpage}
\maketitle

\begin{abstract}
  We argue that grammatical analysis is a viable alternative to
  concept spotting for processing spoken input in a practical spoken
  dialogue system.  We discuss the structure of the grammar, and a
  model for robust parsing which combines linguistic sources of
  information and statistical sources of information.  We discuss test
  results suggesting that grammatical processing allows fast and
  accurate processing of spoken input.
\end{abstract}

\section{Introduction}

\bibliographystyle{fullname}
\label{intro}

The NWO Priority Programme {\em Language and Speech Technology\/} is a
research programme aiming at the development of spoken language
information systems. Its immediate goal is to develop a demonstrator
of a public transport information system, which operates over ordinary
telephone lines.  This demonstrator is
called \ovis, Openbaar Vervoer Informatie Systeem ({\em Public
Transport Information System\/}). The language of the system is Dutch.

At present, a prototype is in operation, which is a version of a 
German system developed by Philips
Dialogue Systems in Aachen \cite{AU95}, adapted to Dutch.
This German system processes spoken input using ``{\em concept
spotting\/}'', which means that the smallest information-carrying
units in the input are extracted,
such as locative phrases (mostly names of train stations) and temporal
expressions, and these are translated more or less individually into updates of the
internal database representing the dialogue state.
The words between the concepts thus perceived are ignored.

The use of concept spotting is common
in spoken-language information systems \cite{WA89,JA91,AU95,AL96}.
Arguments in favour of this kind of shallow parsing are that it is
relatively easy to develop the NLP component, since larger sentence constructs
do not have to be taken into account, and that the robustness of the
parser is enhanced, since sources of ungrammaticality occurring between
concepts are skipped and therefore do not hinder the translation of the
utterance to updates.

The prototype presently under construction (\ovistwo) is based on a
grammar which describes grammatical sentences, i.e. complete and
well-formed user utterances, and thus differs radically from a
concept spotting approach. This article presents a detailed account of
a computational grammar for Dutch, and a robust parsing algorithm
which incorporates this grammatical knowledge as well as other
knowledge sources, such as acoustic evidence and
Ngram statistics. We argue that robust parsing can be based on
sophisticated grammatical analysis. In particular, the grammar
describes full sentences, but in doing so, also describes the grammar
of temporal expressions and locative phrases which are crucial for
concept spotting. Robustness is achieved by taking these phrases into
consideration, if a full parse of an utterance is not available. We
show that our approach is feasible in terms of both accuracy and
computational resources, and thus is a viable alternative to pure
concept spotting.

Whereas some (e.g.\ Moore, Pereira, and Murveit \shortcite{MO89a}) argue that grammatical analysis may
improve recognition accuracy, our current experiments have as yet not
been able to reveal a substantial advantage in this respect.  However, the
grammatical approach may become essential as soon as the application
is extended in such a way that more complicated grammatical
constructions need to be recognized.  In that case, simple concept
spotting may not be able to correctly process all constructions,
whereas the capabilities of the grammatical approach extend much
further.

The structure of this paper is as follows. In section~\ref{grammar} we
describe the grammar for \ovistwo.  We present the grammar in some
detail, since we believe it constitutes an interesting compromise
between linguistic and computational considerations.  Readers
interested in processing issues rather than the details of linguistic
analysis might prefer to skip section~\ref{grammar} (possibly except
the first paragraph) and jump to section~\ref{parser} immediately.
That section describes the robust parsing algorithm.
Section~\ref{eval} reports test results, showing that grammatical
analysis allows fast and accurate processing of spoken input.

\section{A computational grammar for Dutch}
\label{grammar}

In developing the \ovistwo\ grammar we have tried to combine the
short-term goal of developing a grammar which meets the requirements
imposed by the application (i.e.\ robust processing of the output of
the speech recogniser, extensive coverage of locative phrases and
temporal expressions, and the construction of fine-grained semantic
representations) with the long-term goal of developing a general,
computational, grammar which covers all the major constructions of
Dutch.

The design and organisation of the grammar, as well as many aspects of
the particular grammatical analyses we propose, are based on
Head-driven Phrase Structure Grammar \cite{hpsg2}. We depart from this
formalism mostly for computational reasons. As is explained below, the
grammar is compiled into a restricted kind of definite clause grammar
for which efficient processing is feasible. The semantic component
follows the approach to monotonic semantic interpretation using {\em
  quasi-logical forms} presented originally in Alshawi \shortcite{cle-book}.

The grammar currently covers the majority of verbal subcategorisation
types (intransitives, transitives, verbs selecting a {\sc pp}, and
modal and auxiliary verbs), {\sc np}-syntax (including pre- and
post-nominal modification, with the exception of relative clauses),
{\sc pp}-syntax, the distribution of {\sc vp}-modifiers, various
clausal types (declaratives, yes/no and {\sc wh}-questions, and
subordinate clauses), all temporal expressions and locative phrases
relevant to the domain, and various typical spoken-language
constructs.  Due to restrictions imposed by the speech recogniser, the
lexicon is relatively small (3200 word forms, many of which are names
of stations and cities). 

In sections ~\ref{formalism}-~\ref{lexicon} we introduce the grammar
formalism from both a computational and linguistic
perspective. Section~\ref{syntactic_coverage} describes the grammar
of noun, prepositional, and verb phrases, subordinate and main clauses,
{\sc wh}-questions and topicalisation, and a number of domain specific
constructions. Sections~\ref{semantics} and~\ref{updates}, finally, are
concerned with semantics and the translation of quasi-logical forms
into (application-specific) update-expressions. 

\subsection{Formalism}
\label{formalism}
The formalism that we use for the \ovistwo\ Grammar is a variant of
Definite Clause Grammar (DCG) \cite{dcg}. We have chosen for DCG because:

\begin{itemize}
\item DCG provides for a balance between {\em computational
    efficiency} on the one hand and {\em linguistic expressiveness} on
  the other.
\item DCG is a (simple) member of the class of declarative and
  constraint-based grammar formalisms. Such formalisms are widely used
  in linguistic descriptions for NLP. 
\item DCG is straightforwardly related to context-free
  grammar. Almost all parsing technology is developed for CFG;
  extending this technology to DCG is usually possible (although there
  are many non-trivial problems as well). 
\item The compilation of richer constraint-based grammar formalisms
  into DCG is well investigated and forms the basis of several
  wide-coverage and robust grammar systems (i.e. the Alvey-grammar 
  \cite{Alvey,Carroll-diss,Briscoe-Carroll-CL} and the Core Language
  Engine \cite{cle-book}). 
\end{itemize}

\noindent
The formalism for the grammar of \ovistwo\ imposes the following
additional requirements:
\begin{itemize}
\item External Prolog calls (in ordinary DCG these are introduced in
  right-hand sides using curly brackets) are allowed, but must
  be resolved during grammar compilation time. 
\item Rules can be mapped to their `context-free skeleton' (by
  taking the functor symbol of the terms appearing in the right-hand
  and left-hand sides of the rule). This implies that we do not allow the
  under-specification of categories in rules. This is motivated by our
  desire to experiment with parsing strategies in which part of the
  work is achieved on the basis of the context-free skeleton of the
  grammar. It also facilitates indexing techniques.
\item An identifier is assigned to each rule. Such rule identifiers
  have a number of possible uses (debugging, grammar filtering,
  grammar documentation). 
\item The grammar specifies for each rule which daughter 
      is the head. This allows head-driven parsing strategies.
\end{itemize}

\noindent An efficient head-corner parsing strategy for this formalism
is discussed in van Noord \shortcite{cl97}. The restriction that external Prolog
calls must be resolved at compilation time implies that we do not use 
delayed evaluation. More in particular, lexical rules (deriving a
lexical entry from a given `basic' lexical entry) must be applied at
compile time and are not interpreted as (relational) constraints on
under-specified lexical entries, as in van Noord and Bouma 
\shortcite{gj-gosse-kyoto}. Although
we have experimented with combinations of delayed evaluation and
memoisation, as described in Johnson and D\"orre 
\shortcite{johnson-doerre}, the resulting
systems were not efficient enough to be applied in the kind of
practical system considered here. 

\paragraph{Grammar rules.}
A grammar rule is defined by a ternary predicate, {\tt rule/3}. The
first argument of this predicate is a ground Prolog term indicating
the rule identifier. The second argument of the rule is the mother
category. Categories are non-variable Prolog terms.  The third
argument of the rule is a list of categories. Note that we require
that the length of the list is given, and that none of the categories
appearing in the list is a variable. An example of a grammar rule
is provided:
\begin{lnumbered}{test}
\begin{verbatim}
rule(vp_vpnp, vp(Subj,Agr,Sem),
     [v(Subj,Agr,trans,l(Arg,Sem)),np(_,Arg)]).
\end{verbatim}
\end{lnumbered}
\noindent 
Terminal symbols cannot be introduced in rules directly, but
are introduced by means of lexical entries.

\paragraph{Lexical entries.}
The lexicon is defined by the predicate {\tt lex/2}. As an
example, the lexical entry `sleeps' could be encoded as:
\begin{numbered}\begin{verbatim}
lex(sleeps,v(np,agr(3,sg),intrans,l(X,sleep(X)))).
\end{verbatim}\end{numbered}
The first argument is the terminal symbol introduced by this
lexical category.  The second argument is the category (a
non-variable term).  In cases where a lexical entry introduces a
sequence of terminal symbols the first argument is also allowed to
be a (non-empty) list of atoms.

\paragraph{Top category.}
The top category for the grammar (or {\em start symbol\/}) is defined by
the unary predicate {\tt top\_category}. Its argument is an arbitrary
non-variable Prolog term. 

\paragraph{Feature constraints.}
Almost all work in computational grammar writing uses
`feature-structures' of some sort. It is fairly standard to compile
(descriptions of) such features-structures into first-order terms 
(see Pulman \shortcite{Pulman96} for a recent overview).  We use the {\sc hdrug}
development platform \cite{hdrug-paper}, which contains a library for
compiling feature constraints into Prolog terms, and various
predicates to visualise such Prolog terms as feature structures in
matrix notation.

The most important operators provided by the {\sc hdrug} library are the type
assignment operator ('\verb.=>.'), the path equality operator
('\verb.<=>.'), and the path operator ('\verb.:.'). 
A typical grammar fragment employing those operators
is:
\begin{numbered}\begin{verbatim}
rule(1,S,[Np,Vp]) :-
    S => s, np(Np), vp(Vp),
    Vp:vform => finite,
    subj_agreement(Vp,Np).

np(Np) :- Np => np, Np:lex => -.
vp(Vp) :- Vp => v,  Vp:lex => -.
subj_agreement(Vp,Np) :- Vp:agr <=> Np:agr.
\end{verbatim}\end{numbered}
In this rule, the constraint \verb.Np:lex => -. indicates that the
value of the {\tt lex} attribute of {\tt Np} is of type {\tt -}. The
constraint \verb.Vp:agr <=> Np:agr. indicates that the value of the {\tt
  agr} attribute of {\tt Vp} is identical to the value of the {\tt
  agr} attribute of {\tt Np}.
Internally, such a rule could be represented as follows (the actual
result of the compilation depends on what attributes are allowed for
what types; declarations of this sort are part of the grammar):
\begin{numbered}\begin{verbatim}
rule(1,s,[np(Agr,-),v(Agr,-,finite,_,_)]).
\end{verbatim}\end{numbered}
We often will write such rules in matrix notation, as follows: 

\begin{numbered}\begin{avm}
\mbox{\tt rule}(
$ \mbox{1}$,
\mbox{\it s},
$\langle$
\[\avmspan{\it np}\\
   agr&\@1\\
   lex&\mbox{-}
\],

\[\avmspan{\it v}\\
      agr&\@1\\
      lex&\mbox{-}\\
    vform&finite
\]
$\rangle$
).
\end{avm}\end{numbered}

The feature library also supports boolean combinations of atomic values;
these are compiled into Prolog terms using a technique described in
Mellish \shortcite{mellish} (who attributes it to Colmerauer) and
Pulman \shortcite{Pulman96}. Thus, we may specify  {\tt agr} values such as 
{\tt sg $\wedge$ (sec $\vee$ thi)}, denoting an agreement value which
is singular and either second or third person. 

We have also found it useful to provide the predicates {\tt
  unify\_ifdef/3}, {\tt ifdef/4}, and {\tt unify\_except/3}. The predicate {\tt
  unify\_ifdef(C1,C2,Att)} can be used to require that if both {\tt
  C1} and {\tt C2} can have the attribute {\tt Att} (i.e. {\tt C1, C2}
are of a type for which {\tt Att} is a possible feature), then the
values {\tt C1:Att} and {\tt C2:Att} must be identical.  The predicate
{\tt ifdef(Att,Cat,Val,Otherwise)} is used to require that {\tt
  Cat:Att} is identical to {\tt Val} if {\tt Att} is an appropriate
feature for {\tt Cat}. Otherwise {\tt Val} is identical to {\tt
  Otherwise}. The predicate {\tt unify\_except(C1,C2,Path)} unifies {\tt C1}
  and {\tt C2}, with the exception of the value of {\tt Path}, which
  must be defined for both {\tt C1} and {\tt C2}, but which may have 
incompatible values. These predicates simplify the definition of the
grammar code below.

\subsection{Signs}

In unification-based grammar formalisms, linguistic information is
represented by means of typed feature-structures. Each word or phrase
in the grammar is associated with such a feature-structure, in which
syntactic and semantic information is bundled. Within Head-driven
Phrase Structure Grammar ({\sc hpsg}), such feature-structures are
called {\em signs}, a terminology which we will follow here.

At present, the grammar makes use of 15 different types of sign,
where each type roughly corresponds to a different category in
traditional linguistic terminology. For each type of sign, a number of
features are defined. For example, for the type {\sc np}, the features
{\sc agr}, {\sc nform}, {\sc case}, and {\sc sem} are defined. These
features are used to encode the agreement properties of an {\sc np},
(morphological) form, case and semantics, respectively. A more detailed
presentation of these features follows below. 

There are a number of features which occur in most types of sign, and
which play a special role in the grammar. The feature {\sc sc} ({\sc
  subcategorisation}) (present on signs of type {\em v, sbar, det, a, n} and
{\em p}), for instance, is a feature whose value is a list of signs. It
represents the subcategorisation properties of a given sign. As will
be explained below, it is used to implement rules which perform
functor-argument application (as in Categorial Grammar). 

The feature {\sc slash} is present on {\em v, ques} and {\em sbar}.
Its value is a list of signs. It is used to implement a (restricted)
version of the account of nonlocal dependencies proposed in
Pollard and Sag \shortcite{hpsg2} and Sag \shortcite{Sag-Relatives}.  The value of {\sc slash} is
the list of signs which are `missing' from a given constituent. Such a
`missing' element is typically connected to a preposed element in a
topicalisation sentence or {\sc wh}-question.  The same mechanism can
also be used for relative clauses.

The feature {\sc vslash} is similar to {\sc slash} in that it records
the presence of a missing element, a verb in this case. It is used to
implement an account of Dutch main clauses, based on the idea that
main clauses are structurally similar to subordinate clauses, except
for the fact that the finite verb occurs as first or second
constituent within the clause and the clause final position where
finite verbs occur in subordinate clauses is occupied by an {\em
  empty} verbal sign (i.e. an element which is not visible in the
phonological or orthographic representation of the sentence). 

The feature {\sc sem} is present on all signs. It is used to encode
the semantics of a word or phrase, encoded as a {\em quasi
  logical form} \cite{cle-book}.  The feature {\sc mod} is present on
the types {\em a, pp, p, adv, sbar} and {\em modifier}. It is used to
account for the semantics of modifiers. Its value is a list of {\em
  quasi-logical forms}.  In the sections below on syntax, we only give
an informal impression of the semantics. The details of the semantic
construction rules and principles are dealt with in 
section~\ref{semantics}. 

An important restriction imposed by the grammar-parser interface is
that each rule must specify the category of its mother and
daughters. A consequence of this requirement is that
general rule-schemata, as used in Categorial Grammar and {\sc hpsg}
cannot be used in the \ovistwo\ grammar. A rule which specifies that
a head daughter may combine with a complement daughter, if this
complement unifies with the first element on {\sc sc} of the head
(i.e. a version of the categorial rule for functor-argument
application) cannot be implemented directly, as it leaves the
categories of the daughters and mother unspecified. Nevertheless,
generalisations of this type do play a role in the grammar. We adopt
an architecture for grammar rules similar to that of {\sc hpsg}, in
which individual rules are classified in various {\em structures},
which are in turn defined in terms of general {\em principles}. 

Rules normally introduce a structure in which one of the daughters can
be identified as the {\em head}. The head daughter either
subcategorises for the other (complement) daughters or else is
modified by the other (modifier) daughters.

The two most common structures are the {\em head-complement} and {\em
  head-modifier} structure.\footnote{Other structures are the {\em
    main-clause} and {\em head-filler} structure.  These are discussed
  in the sections on main-clause syntax and topicalisation. } In
figure~\ref{principles} we list the definitions for these structures
and the principles they refer to, except for the {\em filler}
principle, which is presented in the section on topicalisation.

\begin{figure}
\begin{verbatim}
hd_comp_struct(Head,Complements,Mother) :-
    hd_struct(Head,Complements,Head,Mother).

hd_mod_struct(Head,Modifier,Mother) :-
    hd_struct(Head,[],Modifier,Mother),
    Head:sem <=> HeadSem,
    Modifier:mod <=> [HeadSem].

hd_struct(Head,Complements,SemanticHead,Mother) :-
    head_feature_principle(Head,Mother),
    valence_principle(Head,Complements,Mother),
    filler_principle(Head,[],Mother),
    SemanticHead:sem <=> Mother:sem.
      
head_feature_principle(Head,Mother) :-
    unify_ifdef(Head,Mother,vform),
    unify_ifdef(Head,Mother,agr),
    unify_ifdef(Head,Mother,case),
    unify_ifdef(Head,Mother,mod),
    unify_ifdef(Head,Mother,pform),
    unify_ifdef(Head,Mother,aform),
    unify_ifdef(Head,Mother,vslash),
    unify_ifdef(Head,Mother,subj).

valence_principle(Head,Complements,Mother) :-
    ifdef(sc,Head,HeadSc,[]),
    ifdef(sc,Mother,MotherSc,[]),
    append(Complements,MotherSc,HeadSc) 
\end{verbatim}
\caption{Structures and Principles}
\label{principles}
\end{figure}

{\em Head-complement} and {\em head-modifier} structures are instances
of  {\em headed} structures. The definition of {\em headed} structure
refers to the {\sc head-feature}, {\sc valence}, and {\sc filler}
principles, and furthermore fixes the semantic head of a phrase. Note
that the definition of {\tt hd-struct} has a number of parameters. The
idea is that a headed structure will generally consist of a head
daughter, and furthermore of zero or more complement daughters and possibly
a modifier. {\em Head-complement} and {\em head-modifier} structures
differ from each other only in that the first introduces complements,
but no modifiers, whereas the second introduces no complements, but a
modifier. Moreover, the syntactic head is also the semantic
head in {\em head-complement} structures, but not in a {\em
  head-modifier} structure.  In {\em head-modifier} structures, the
semantic contribution of the head to the meaning of the phrase as a
whole is handled by unifying the head semantics with the value of (the
first element of) {\sc mod} on the modifier.

The {\sc head feature principle} states for a number of features (the
{\em head-features}) that their value on the head daughter
and mother must be unified. As this principle generalises over various
types of sign, its definition requires the predicate {\tt
  unify\_ifdef}.  

The {\sc valence principle} determines the value of the {\em
  valence feature} {\sc sc}. The value of
{\sc sc} on the head daughter of a rule
is the concatenation ({\tt append}) of the list of complement
daughters and the value of {\sc sc} on the mother. Another way to put
this is that the value of {\sc sc} on the mother is the value of {\sc
  sc} on the head daughter minus the elements on {\sc sc} that
correspond to the complement daughters. Note that the formulation of
the {\sc valence principle} is complicated by the fact that {\sc sc}
(or {\sc subj}) may sometimes not be defined on the mother. In that
case, it is assumed that the value of {\sc sc} on the head daughter
must correspond exactly to the list of complement daughters. The
constraint {\tt ifdef(sc,Mother,MotherSc,[])} states that the value of
{\sc sc} on {\tt Mother} unifies with {\tt MotherSc}, if {\sc sc} is
defined for the type of {\tt Mother}. Otherwise, {\tt MotherSc} is
assigned the value {\tt []} (i.e. the empty list).

The structures defined in figure~\ref{principles} are used in the
definition of grammar rules. The {\tt np-det-n} rule introduces a {\em
  head-complement} structure in which (following the traditional semantic
analysis) the determiner is the head, and
the noun the complement: 
\begin{lnumbered}{np-rules-a}\begin{verbatim}
rule(np_det_n, NP, [Det, N]) :-    
    NP => np, Det => det, N => n,
    NP:nform => norm, hd_comp_struct(Det,[N],NP).
\end{verbatim}\end{lnumbered}
The {\tt n-adj-n} rules introduces a {\em
  head-modifier} structure where the adjective is the modifier: 
\begin{lnumbered}{np-rules-b}\begin{verbatim}
rule(n_adj_n, N1, [AdjP, N0]) :-
    N1 => n, AdjP => a, N0 => n,
    AdjP:agr <=> N0:agr, hd_mod_struct(N0,AdjP,N1).
\end{verbatim}\end{lnumbered}

Note
that for a given rule, the types of the mother and daughters must be
specified, and furthermore, the number of complements is always
specified. This implies that the constraints in the principles in 
figure~\ref{principles} can be reduced to a number of basic
constraints on the values of particular features defined for the signs
in the rule.  The previous two rules 
can be depicted in matrix notation as (where  $\langle\rangle$ denotes
the empty list):
\begin{numbered}
\begin{avm}
{\tt np\_det\_n:}
\[\avmspan{\it np}\\agr  & \@1 \\ nform & norm \\ sem & \@2 \] ~~ $\rightarrow$ ~
   \[\avmspan{\it det}\\ sc & \< \@3  \> \\ agr & \@1 \\ sem & \@2 \]~~~
   \@3 n
\end{avm}
\end{numbered}

\begin{numbered}
\begin{avm}
{\tt n\_adj\_n:}
\[\avmspan{\it n}\\sc & \@1 \\ agr  & \@2 \\ sem & \@3 \] ~~ $\rightarrow$ ~
   \[\avmspan{\it a}\\ sc & $\langle\rangle$ \\ agr & \@2 \\ sem & \@3  \\ mod &  \< \@4 \> \]~~~
   \[\avmspan{\it n}\\sc & \@1 \\ agr  & \@2 \\ sem & \@4 \]
\end{avm}
\end{numbered}

\begin{figure}
\includegraphics[scale=0.6]{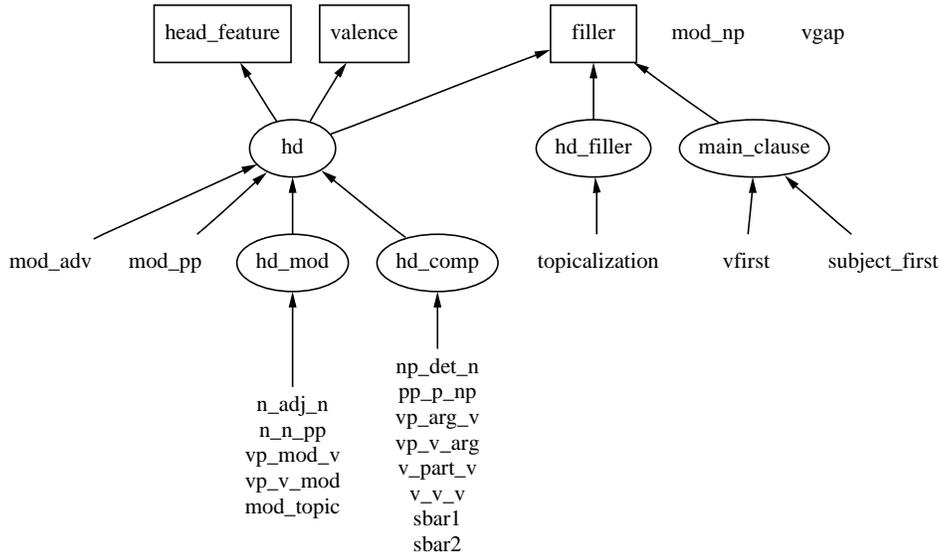}
\centering
\caption{The Rule Hierarchy (with {\sc Principles} shown in boxes, {\em
    structures} in ovals, and {\tt rules} without frame). Note that
the {\tt mod\_np} rule (a unary rule which transforms temporal {\sc np}s into
verbal modifiers) and the {\tt vgap} rule (a rule which introduces
verbal gaps) are exceptional in that they do not inherit from general
principles.}
\label{rules}

\end{figure}

An overview of all grammar rules defined in the fragment at the
moment, together with the structures and principles from which they
inherit, is given in figure~\ref{rules}.

The classification of rules into structures, which are in turn defined
in terms of principles, allows us to state complicated rules
succinctly and to express a number of generalizations. Nevertheless,
it is also clear that the rules could have been more general, if rule
{\em schemata} (in which the type of the daughters, or even the number
of daughters is not necessarily specified) had been allowed.  Given
this restriction, one may even wonder whether the {\sc valence
  principle} (and the feature {\sc sc} that comes with
it) cannot be eliminated in favour of more specific rules. Valence
features are particularly important for grammars employing rule
schemata, but they are much less crucial for more traditional types of
grammar.  Although eliminating valence features is not impossible in
principle, we believe that the present set-up still has advantages,
although these are less apparent than in grammars which make use of
rule schemata.  Expressing valence information lexically, instead of
using more detailed syntactic rules, has the advantage that
idiosyncratic subcategorization requirements (such as the restriction
that {\em denken} ({\em to think}) requires a {\sc pp}-complement
headed by {\em aan} ({\em about}), or the fact that {\em komen} ({\em
  to come}) may combine with the particle {\em aan} (the combination
of which means {\em to arrive})) need not be stated in the rules.
Similarly, all constraints having to do with case marking and
agreement can be expressed lexically, as well as the semantic relation
between a head and its dependents.

\subsection{The lexicon}
\label{lexicon}

The lexicon is a list of clauses {\tt lex(Word,Sign)},
associating a word (or sequence of words) with a specific sign.

Constraint-based grammars in general, and lexicalist constraint-based
grammars in particular, tend to store lots of grammatical information
in the lexicon. This is also true for the \ovistwo\ grammar. A lexical
entry for a transitive verb, for instance, not only contains
information about the morphological form of this verb, but also
contains the features {\sc sc} and {\sc subj} for which quite detailed
constraints may be defined. Furthermore, for all lexical signs it is
the case that their semantics is represented by means of a
feature-structure. This structure can also be quite complex. To avoid
massive reduplication of identical information in the lexicon, the use
of inheritance is therefore essential.

\begin{figure}
\begin{verbatim}
intransitive(Pred,Sign) :- iv(Sign), iv_sem(Sign,Pred).

transitive(Pred,Sign) :-   tv(Sign), tv_sem(Sign,Pred).

v(V) :- V => v, V:lex => basic, 
    V:vslash => [], V:subj <=> [Subj], 
    Subj => np, Subj:nform => norm.

iv(V) :- v(V), V:sc <=> [].

tv(V) :- v(V), V:sc <=> [Obj],
    Obj => np, Obj:nform => norm, Obj:case => acc.

weather_v(V) :- iv(IV), unify_except(IV,V,subj:h:nform),
    V:subj:h:nform => it.
\end{verbatim}
\caption{Fragment of the lexical hierarchy}
\label{lex-hier}
\end{figure}

In figure~\ref{lex-hier}, we illustrate the use of inheritance in the
lexicon. All lexical entries for verbs have a number of properties in
common, such as the fact that they are of type {\em v}, and take a
normal (non-locative and non-temporal) {\sc np} as subject.
This is expressed by the template {\tt v(V)}.
Intransitive verbs ({\tt iv(V)}) can
now be characterised syntactically as verbs which do not subcategorise
for any (non-subject) complements. Transitive verbs ({\tt tv(V)})
subcategorise for an {\sc np} with accusative case. The templates {\tt
  intransitive(Pred,Sign)} and {\tt transitive(Pred,Sign)}, finally,
combine the syntactic and semantic properties of intransitive and
transitive verbs. The variable {\tt Pred} is used in the semantics to
fix the value of the predicate defined by a particular verb. A limited
form of non-monotonic inheritance is supported (see Carpenter
\shortcite{Carpenter-defaults}
and Bouma \shortcite{bouma-cl} for more general approaches). For instance, 
`weather' verbs require the dummy pronoun {\em het} ({\em it}) as
subject, but behave otherwise as intransitive verbs. This can be
expressed by letting {\tt weather\_v} inherit from {\tt iv}, with the
exception of the value of the {\sc nform} attribute of (head of the
list containing) the subject, which is assigned an exceptional value. 
The attribute-value matrices for the templates {\tt iv(V)} and {\tt tv(V)}
are: 

\begin{numbered}\begin{avm}
\mbox{\tt iv}(\[\avmspan{\it v}\\lex & basic \\ sc & $\langle\rangle$ \\ 
              subj & \< \[\avmspan{\it np}\\ nform & norm \] \> \\
              vslash & $\langle\rangle$
              \]~).
\end{avm}~~~\begin{avm}
\mbox{\tt tv}(\[\avmspan{\it v}\\lex & basic \\ sc & \< \[\avmspan{\it np}\\nform & norm \\
                                                  case & acc \]\> \\ 
              subj & \< \[\avmspan{\it np}\\nform & norm \] \> \\
              vslash & $\langle\rangle$
              \]~).
\end{avm}\end{numbered}

The lexicon itself (i.e. the predicate {\tt lex/2}) is defined in
terms of the predicates {\tt entry}, {\tt inflection} and {\tt
  lexical\_rules}:  
\begin{numbered}\begin{verbatim}
lex(Word,Sign) :-
    entry(Root,Sign0),
    inflection(Root,Word,Sign0,Sign1),
    lexical_rules(Sign1,Sign).
\end{verbatim}\end{numbered}
The definition of {\tt entry(Root, Sign)} defines for each root
form what its associated sign is. For instance, for verbs we must
typically distinguish a first person singular form, a second and
third person singular form, and a plural form (which is also the
form of the infinitive). The predicate {\tt inflection} defines
how inflected forms are derived. For example, there is an
inflection rule which adds a {\tt t} to the base form of a verb,
and specifies that its agreement features are third person
singular, and its {\sc vform} value is {\it fin}.  Lexical rules
can be used to transform the sign associated with a lexical entry.
For instance, the account of nonlocal dependencies sketched below
makes use of a lexical rule which removes a sign from {\sc sc} and
places it on {\sc slash}. A more detailed account of this lexical
rule is given in the section on nonlocal dependencies.
As an example, assume the stem {\tt arriveer} (to arrive) is defined as an
intransitive:
\begin{numbered}\begin{verbatim}
entry(arriveer,Sign):-
    intransitive(arriveren,Sign).
\end{verbatim}\end{numbered}    
Such a definition will give rise to a number of lexical entries. One
of these will be the third person singular finite form:
\begin{numbered}\begin{avm}
lex(arriveert,\[\avmspan{\it v}\\lex & basic \\ vform & fin \\ sc & $\langle\rangle$ \\ 
              subj & \< \[\avmspan{\it np}\\ nform & norm \\ 
                                     agr & sg $\wedge$ thi\] \> \\
              vslash & $\langle\rangle$
              \]~).
\end{avm}\end{numbered}

\subsection{Syntactic Coverage}
\label{syntactic_coverage}

Below, we describe the syntactic coverage of the grammar. The grammar
is not intended as a general, wide-coverage, grammar for Dutch. This
implies not only that coverage in the lexical domain is limited, but
also that several grammatical constructions are not taken into
consideration (e.g. passives) or accounted for only to a certain
extent (e.g. the grammar of Dutch verb clusters). The coverage of the
grammar is quite satisfactory for the \ovis\ application, however. For
instance, when evaluating  the grammar on a corpus of 1000 transcribed
test-sentences, we obtained a semantic concept accuracy of 95\% (see section 4.2
for discussion). 

\subsubsection{Noun phrases}
\label{nps}

The four types which are relevant in the syntax of noun phrases are
{\it np} (noun phrase), {\it det} (determiner), {\it a} (adjective)
and {\it n} (noun). Each type has the attributes {\sc agr} and {\sc
  sem}. Furthermore, {\it det} and {\it n} have an attribute {\sc
  sc}. The {\it np} type has three further attributes: {\sc case}, 
{\sc nform} and {\sc pform}. Finally, {\it a} is also specified for {\sc mod}. 

The features {\sc agr} ({\em agreement}), {\sc case}, and {\sc nform}
({\em noun form}) are used to encode agreement properties (encoded as
a boolean combination of {\em person}, {\em number}, {\em determiner}
and {\em definiteness}), the case value and the form of an item. Their
possible values are listed in (\ref{agr}). Note that {\sc agr}
contains the information needed for subject-verb agreement, as well as
for {\sc np}-internal agreement (between determiner, adjective, and
noun). The agreement types {\em de} and {\em het} (the two forms of
the definite article) distinguish between neuter and nonneuter nouns.
{\sc Case} and {\sc nform} are relevant for full {\sc np}s only. 
\begin{lnumbered}{agr}
\begin{tabular}[t]{ll}
{\em Agr} &  (fir $\vee$ sec $\vee$ thi) $\wedge$ 
             (sg $\vee$ plu) $\wedge$ (de $\vee$ het) $\wedge$
             (def $\vee$ indef) \\
{\em Case} & nom $\vee$ acc \\
{\em Nform} & norm $\vee$ loc $\vee$ temp $\vee$ num
\end{tabular}
\end{lnumbered}

Full {\sc np}s never take complements, so they do not have a
feature {\sc sc}.  Adjectives may modify a noun, therefore the feature
{\sc mod} is defined for type {\em a}.

\begin{figure}
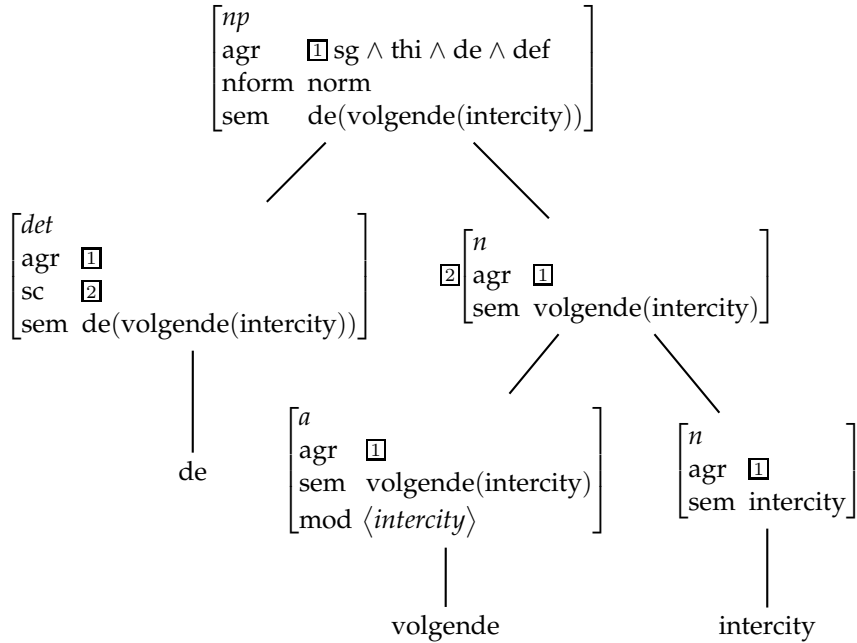

\pstree[levelsep=*1cm,nodesep=3pt]{\Tr[ref=c]{
\begin{avm}\[\avmspan{\it np}\\agr & \@1 sg $\wedge$ thi $\wedge$ de  $\wedge$ def \\
            nform & norm \\
            sem & de$($volgende$($intercity$))$ \]\end{avm}}}{
  \pstree{\Tr[ref=c]{\begin{avm}\[\avmspan{\it det}\\agr & \@1 \\
             sc & \@2 \\
             sem & de$($volgende$($intercity$))$ \]\end{avm} }}{\Tr[ref=c]{\mbox{de}}}
  \pstree{\Tr[ref=c]{         \begin{avm}\@2\[\avmspan{\it n}\\agr & \@1 \\
              sem & volgende$($intercity$)$ \]\end{avm}
}}{
    \pstree{\Tr[ref=c]{  \begin{avm}\[\avmspan{\it a}\\agr & \@1 \\
               sem & volgende$($intercity$)$ \\
               mod & \< {\em intercity} \> \]\end{avm}
}}{\Tr[ref=c]{\mbox{volgende}}}
    \pstree{\Tr[ref=c]{  \begin{avm}\[\avmspan{\it n}\\agr & \@1 \\
               sem &  intercity \]\end{avm} 
}}{\Tr[ref=c]{\mbox{intercity}}}
  }
}
\centering
\caption{de volgende intercity (the next intercity)}
\label{np-deriv}
\end{figure}

The two rules we presented in (\ref{np-rules-a}) and
(\ref{np-rules-b}) (section \ref{syntactic_coverage}) are used to
form {\sc np}s consisting of a determiner and a (possibly complex)
noun, and {\sc n}s consisting of an adjective followed by a (possibly
complex) noun. The derivation of the {\sc np} 
{\em de volgende intercity} (the next intercity) is shown in figure~\ref{np-deriv}.

\subsubsection{Prepositional phrases}

Prepositional phrases are of type {\em pp} and are headed by
prepositions, i.e. elements of type {\em p}. 
Prepositions subcategorize (usually) for an {\sc np}, so the value of
{\sc sc} on {\sc p} will be a list of length one, containing the {\sc
  np}-complement. The feature {\sc pform} takes as value the specific
form of the preposition heading the {\sc pp} (i.e. {\em van, op, naar,
  \ldots}). This information can be used to let a verb select a {\sc
  pp} headed by a specific preposition.  

Full {\sc pp}s can modify nouns or verb phrases. Therefore, {\sc pp}
has a feature {\sc mod}.  {\sc mod} has to be present on
{\sc p} as well, as the relation between the semantics of the
preposition and the element it modifies is encoded as part of the
lexical entry of a preposition.
Here, we give the rule which
forms {\sc pp}s and the rule which lets a {\sc pp} combine as a
modifier with a noun. 

\begin{numbered}\begin{avm}
{\tt pp\_p\_np:}
\[\avmspan{\it pp}\\pform & \@1 \\ sem & \@2 \\ mod & \@3 \] ~~$\rightarrow$~
   \[\avmspan{\it p}\\ pform & \@1 \\ sc & \< \@4  \> \\ sem & \@2 \\ mod & \@3 \]~~~
   \@4 np
\end{avm}\end{numbered}

\begin{figure}
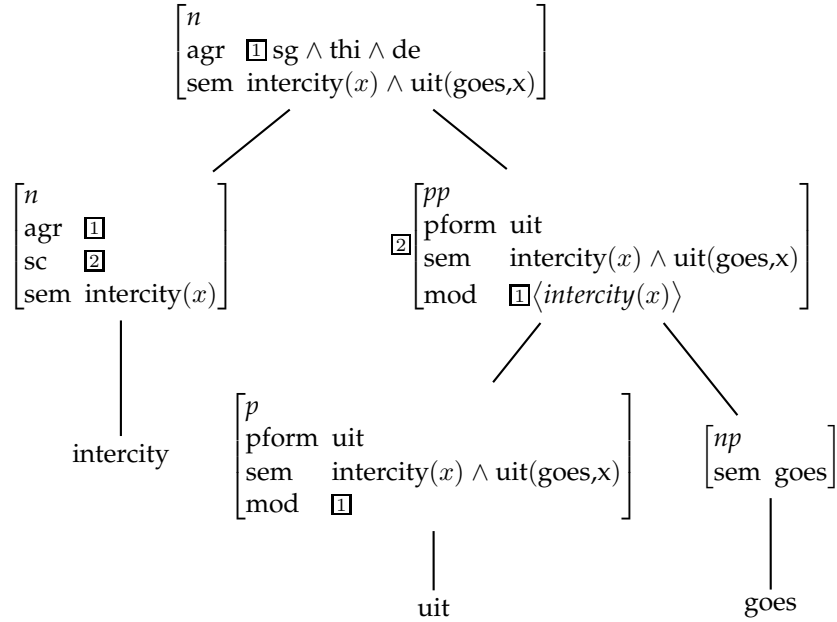

\pstree[levelsep=*1cm,nodesep=3pt]{\Tr[ref=c]{
  \begin{avm}\[\avmspan{\it n}\\agr & \@1 sg $\wedge$ thi $\wedge$ de  \\
              sem & intercity$(x)$ $\wedge$ uit$($goes,x$)$  \]\end{avm}
}}{
  \pstree{\Tr[ref=c]{  \begin{avm}\[\avmspan{\it n}\\agr & \@1 \\
             sc & \@2 \\
             sem & intercity$(x)$  \]\end{avm}
 }}{\Tr[ref=c]{\mbox{intercity}}}
  \pstree{\Tr[ref=c]{          \begin{avm}\@2\[\avmspan{\it pp}\\ pform & uit \\
              sem & intercity$(x)$ $\wedge$ uit$($goes,x$)$ \\
              mod & \@1 \<  {\em intercity$(x)$} \> \]\end{avm}
}}{
    \pstree{\Tr[ref=c]{  \begin{avm}\[\avmspan{\it p}\\ pform & uit \\
               sem & intercity$(x)$ $\wedge$ uit$($goes,x$)$ \\
               mod & \@1 \]\end{avm}
 }}{\Tr[ref=c]{\mbox{uit}}}
    \pstree{\Tr[ref=c]{  \begin{avm}\[\avmspan{\it np}\\
               sem &  goes \]\end{avm} 
 }}{\Tr[ref=c]{\mbox{goes}}}
  }
}
\centering
\caption{intercity uit Goes (intercity from Goes)}
\label{pp-deriv}
\end{figure}

\begin{numbered}
\begin{avm}
{\tt n\_n\_pp:}
\[\avmspan{\it n}\\sc & \@1 \\ agr  & \@2 \\ sem & \@3 \] ~~ $\rightarrow$ ~
   \[\avmspan{\it n}\\sc & \@1 \\ agr  & \@2 \\ sem & \@4 \] ~~~
   \[\avmspan{\it pp}\\ sem & \@3  \\ mod &  \< \@4 \> \]
\end{avm}
\end{numbered}
\noindent Using these rules, we can derive the phrase 
{\em intercity uit Goes} (intercity from Goes) as illustrated in figure~\ref{pp-deriv}.

It should be noted that since adjectives precede the nouns they modify
and {\sc pp}s follow them, an expression such as {\em volgende
  intercity uit Groningen} (next intercity from Groningen) will
receive two parses. This appears to be a case of spurious ambiguity.
There are intensional adjectives, such as {\em zogenaamde} ({\em
  alleged}), which need to be able to take scope over a complex noun,
but it seems that modifying {\sc pp}s never need to take scope over a
adjective + noun combination. It is not easy to rule out the latter
type of derivation, however, without introducing additional features.

\subsubsection{Verb phrases}
\label{vps}

Both verbs and verb phrases are of type {\em v}:
\begin{lnumbered}{v-type}
\begin{avm}
\[\avmspan{\it v}\\ lex & ylex $\vee$ nlex \\ 
            null & null $\vee$ nonnull \\
            vform & fin $\vee$ inf $\vee$ te $\vee$ psp  \\ 
            sc  & listof(Sign) \\
            subj & listof(Sign) \\
            sem & Qlf \\
            slash & listof(Sign) \\
            vslash & Vslash
          \]
\end{avm} 
\end{lnumbered}

The features {\sc lex}, {\sc null}, and {\sc vform} are specific for
{\em v}.  The feature {\sc vform} is used to distinguish finite,
infinitive, {\em te}-infinitive and past participle verbs (and verb
phrases headed by such verbs). The feature {\sc lex} is used to
distinguish lexical verbs ({\em ylex}) from verbal phrases that are
not lexical ({\em nlex}). The feature $ylex$ subsumes two further
subtypes {\sf basic} $\vee$ {\sf complex}, to distinguish basic and
complex lexical verbs. The latter are combinations of a verb and a
separable prefix ({\em aan+komen}, arrive) or combinations of a modal verb and
a main verb ({\em wil vertrekken}, want to leave). The feature {\sc null} is used to
distinguish verbal {\em traces} (i.e. verbal signs without
phonological content) from other verbal signs.  The features {\sc
  subj}, {\sc slash}, and {\sc vslash} and {\sc null} are discussed in
the section below on sentential syntax.

There are a number of similar rules for combining a verb or a verbal
projection with one of its complements. One rule combines a noun
phrase complement with a verbal head ({\em een kaartje kopen}, buy a ticket): 
\begin{numbered}\begin{avm}
{\tt vp\_np\_v:} 
\[\avmspan{\it v}\\lex & nlex \\ sc & \@1 \\ vform & \@2 \\ slash & \@3 \\
           vslash & \@4 \] ~~$\rightarrow$~
\@6 \sort{np}{}~~~
\[\avmspan{\it v}\\ sc & \<\@6$|$\@1\> \\ vform & \@2 \\ slash & \@3 \\
           vslash & \@4 \]
\end{avm}\end{numbered}
Since {\sc pp}s may either precede or follow the head ({\em vanuit
  Leiden vertrekken, vertrekken vanuit Leiden}, depart from Leiden),
there are two rules to combine such a {\sc pp} and a verbal head.
Finally, there is a rule which combines a verbal head with a {\em
  te}-infinitive ({\em weigeren naar Groningen te komen}, refuse to
come to Groningen).  The result of combining a verb (or verbal
projection) with its complement is a phrase (i.e. the value of {\sc
  lex} on the mother is {\em nlex}).

A verbal modifier can be either an adverb, a {\sc
  pp}, or a temporal {\sc np}. There are unary rules rewriting signs
of type modifier into each of these categories. One such rule is the
following: 
\begin{numbered}\begin{avm}
{\tt mod\_adv:} 
\[ \avmspan{\it modifier}\\mod & \@1 \< \_ \> \] ~~$\rightarrow$~

\[\avmspan{\it adv}\\mod & \@1 \]
\end{avm}
\end{numbered}

At the moment, we allow all modifiers to precede or follow the verb
({\em ik moet morgen in Assen zijn/ in Assen zijn morgen/ morgen zijn
  in Assen}, I must be in Assen tomorrow, {\em ik moet tien uur in Assen
zijn/?in Assen zijn tien uur}, I must be in Assen at ten o'clock).
Therefore, there are two similar rules, {\tt vp\_v\_mod} and {\tt
  vp\_mod\_v}, in which a verb combines with a modifier. The first is
illustrated here:

\begin{numbered}
\begin{avm}
{\tt vp\_mod\_v:} 
\[\avmspan{\it v}\\lex & nlex \\ sc & \@1 \\ vform & \@2 \\ slash & \@3 \\
           vslash & \@4 \] ~~$\rightarrow$~
\[\avmspan{\it modifier}\\mod & \< \@5 \> \]~~~
\[\avmspan{\it v}\\ sc & \@1 \\ vform & \@2 \\ slash & \@3 \\
           vslash & \@4 \\ sem & \@5 \]
\end{avm}
\end{numbered}

\begin{figure}
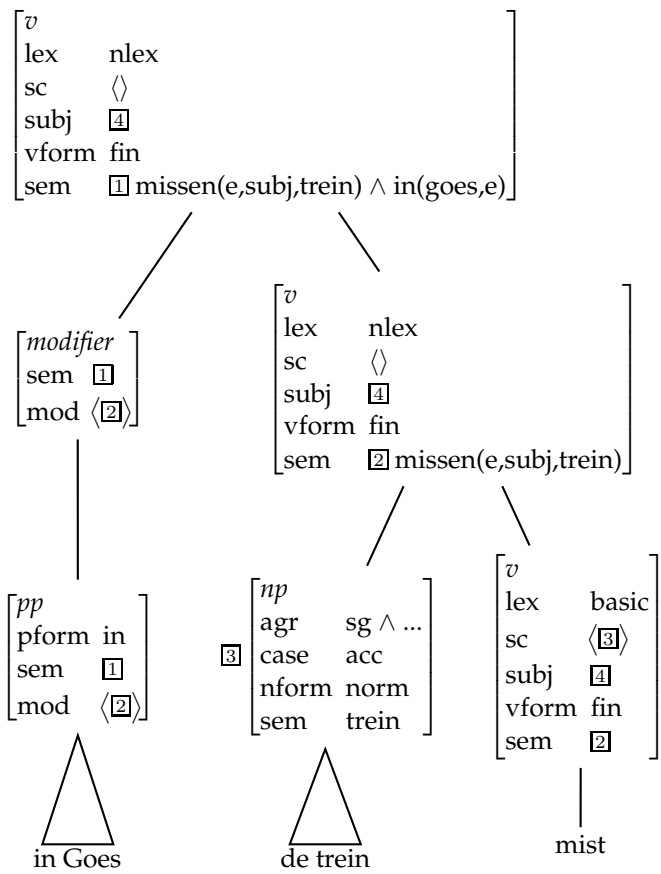

\pstree[levelsep=*1cm,nodesep=3pt]{\Tr[ref=c]{\begin{avm}
    \[\avmspan{\it v}\\ lex & nlex \\ sc & $\langle\rangle$ \\ subj & \@4 \\ vform & fin \\
                sem & \@1 missen(e,subj,trein) $\wedge$ in(goes,e) \]
    \end{avm}}}{
\pstree{\Tr[ref=c]{\begin{avm}
  \[\avmspan{\it modifier}\\ sem & \@1 \\
                     mod & \< \@2 \> \]
\end{avm}}}{
  \pstree{\Tr[ref=c]{ \begin{avm}
  \[\avmspan{\it pp}\\ pform & in \\ 
               sem & \@1 \\
               mod & \< \@2 \> \]
\end{avm}
  }}{\pstree[thislevelsep=5pt,edge=none]{\Tfan}{\pstree{\Tr[ref=c]{in Goes}}{}}}}
\pstree{\Tr[ref=c]{\begin{avm}
  \[\avmspan{\it v}\\ lex & nlex \\ sc & $\langle\rangle$ \\ subj & \@4 \\ vform & fin \\
                sem & \@2 missen(e,subj,trein) \]
\end{avm}}}{
  \pstree{\Tr[ref=c]{  \begin{avm}
  \@3 \[\avmspan{\it np}\\ agr & sg $\wedge$ ... \\
               case & acc \\ 
               nform & norm \\
               sem & trein \]
\end{avm}}}{
       \pstree[thislevelsep=5pt,edge=none]{\Tfan}{\pstree{\Tr[ref=c]{de trein}}{}}}
  \pstree{\Tr[ref=c]{\begin{avm}
  \[\avmspan{\it v}\\   lex & basic \\ sc & \< \@3 \> \\ subj & \@4 \\ vform & fin \\
                sem & \@2 \]
\end{avm} }}{
    \pstree{\Tr[ref=c]{mist}}{}}}
 }
\centering

\caption{(dat Rob) in Goes de trein mist (that Rob misses the train in
  Goes)}
\label{vp-ex1}
\end{figure}

A special type {\em modifier} (with {\sc sem} and {\sc mod} as only
attributes) in combination with three unary rules is
used to introduce the various types of verbal modifier. 
A sample derivation is given in figure~\ref{vp-ex1} (the value of the
features {\sc slash} and {\sc vslash} is not shown, but is
$\langle$~$\rangle$ on all verbal signs in this derivation).

Finally, there are two {\sc vp}-rules that give rise to `complex'
lexical expressions, instead of phrases. Firstly, consider the {\tt
  v\_v\_v} rule:
\begin{numbered}\begin{avm}
{\tt v\_v\_v:}
\[\avmspan{\it v}\\lex & complex \\ sc & \@1 \\ vform & \@2 \\ slash & \@3 \\
           vslash & \@4 \\ sem & \@5 \] ~~$\rightarrow$~

\[\avmspan{\it v}\\ lex & basic \\ sc & \<\@6 $|$ \@1\> \\ vform & \@2 
            \\ slash & \@3 \\ vslash & \@4 \\ sem & \@5 \]~~~
\@6 \[\avmspan{\it v}\\ lex & ylex \\ vslash & $\langle\rangle$ \]
\end{avm}\end{numbered}
The {\tt v\_v\_v} rule is used to derive phrases in which a modal verb
precedes its infinitival complement ({\em (dat ik om tien uur)
wil vertrekken}, that I want to leave at ten o'clock). We adopt an
analysis
 of such constructions
in which modals {\em inherit} the arguments on {\sc sc} of the
infinitival verb with which they combine. This is illustrated for the
root {\em wil} (want) in (\ref{willen}). 
\enumsentence{
\label{willen}
\begin{avm}
{\tt wil} ~~$\mapsto$~~
\[\avmspan{\it v}\\ lex & basic \\ 
            sc & \< \[\avmspan{\it v}\\ lex & ylex \\ sc & \@1 \\
                              vform & inf \]$|$ \@1 \> \\
         \]
\end{avm}
}

\begin{figure}
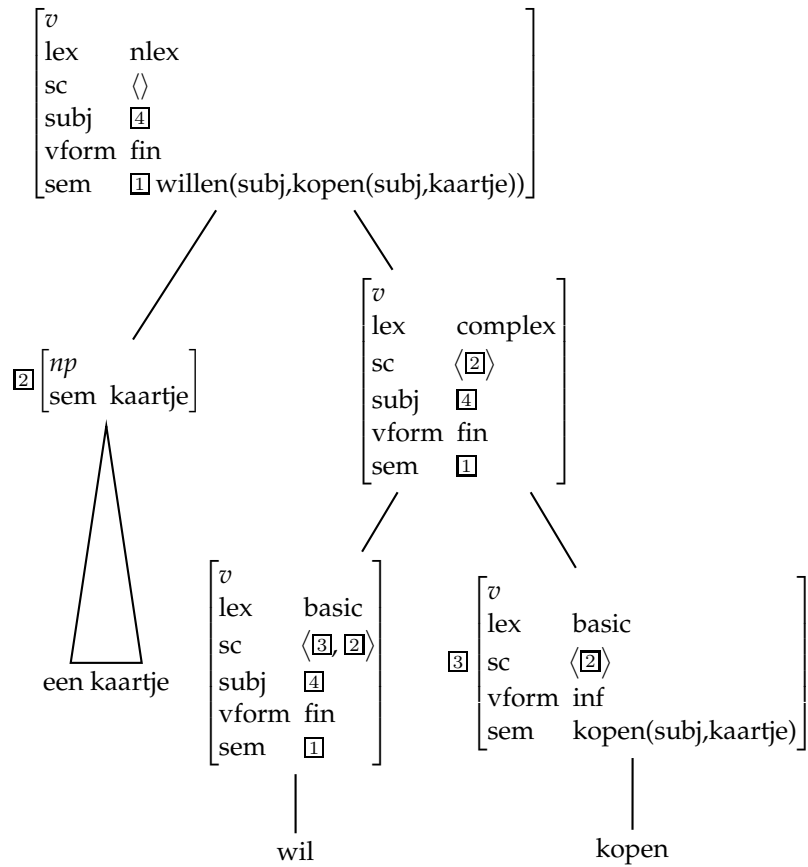

\pstree[levelsep=*1cm,nodesep=3pt]{\Tr[ref=c]{
  \begin{avm}
    \[\avmspan{\it v}\\ lex & nlex \\ sc & $\langle\rangle$ \\ subj & \@4 \\ vform & fin \\
                sem & \@1 willen(subj,kopen(subj,kaartje))  \]
    \end{avm}
 }}{
\pstree{\Tr[ref=c]{  \begin{avm}
  \@2 \[\avmspan{\it np}\\ sem & kaartje \]
\end{avm}
}}{
\pstree[thislevelsep=8pt,edge=none]{\Tfan}{\pstree{\Tr[ref=c]{een kaartje}}{}}
}
\pstree{\Tr[ref=c]{  \begin{avm}
  \[\avmspan{\it v}\\ lex & complex \\ sc & \<\@2 \> \\ subj & \@4 \\ vform & fin \\
                sem & \@1  \]
\end{avm}
}}{
  \pstree{\Tr[ref=c]{  \begin{avm}
  \[\avmspan{\it v}\\ lex & basic \\ sc & \<\@3, \@2 \> \\ subj & \@4 \\ vform & fin \\
                sem & \@1  \]
\end{avm}
}}{
    \pstree{\Tr[ref=c]{wil}}{}}
  \pstree{\Tr[ref=c]{\begin{avm}
  \@3 \[\avmspan{\it v}\\   lex & basic \\ sc & \< \@2 \>  \\ vform & inf \\
                sem & kopen(subj,kaartje) \]
\end{avm} 
}}{
    \pstree{\Tr[ref=c]{kopen}}{}}}
}
\centering
\caption{(dat ik) een kaartje wil kopen (that I want to buy a ticket)}
\label{vp-ex2}
\end{figure}

\noindent 
This allows us to derive phrases
such as {\em (dat ik) een kaartje wil kopen} (that I want to buy a
ticket) where the finite 
modal verb combines with the infinitival verb before combining with
the object of {\em kopen} (figure~\ref{vp-ex2}). Note that it is
essential that the modal verb selects a {\sc [lex  {\em ylex}]} argument in
this case, as this excludes the derivation of ungrammatical
expressions such as {\em (dat ik) wil een kaartje kopen}. The result
of combining a modal with an infinitival verb is {\sc [lex {\em
    complex}]} (i.e. subsumed by {\sc [lex  {\em ylex}]}). This
implies that such combinations can be selected by another modal verb
(i.e. {\em (dat ik) een kaartje zou willen kopen}, that I would like
to buy a ticket). 

Next, consider the {\tt v\_part\_v} rule: 
\begin{numbered}\begin{avm}
{\tt v\_part\_v:}
\[\avmspan{\it v}\\lex & complex \\ sc & \@1 \\ vform & \@2 \\ slash & \@3 \\
           vslash & \@4 \\ sem & \@5 \] ~~$\rightarrow$~
\@6 \sort{part}{}~~~
\[\avmspan{\it v}\\ lex & ylex \\ sc & \<\@6 $|$ \@1\> \\ vform & \@2 
            \\ slash & \@3 \\ vslash & \@4 \\ sem & \@5 \]
\end{avm}\end{numbered}
The rule {\tt v\_part\_v} is used to account for constructions such as 
{\em (dat ik voor tien uur) aan wil komen} (that I want to arrive
before ten o'clock). The prefix (or particle)
{\em aan} of the verb {\em aankomen} (arrive) is separated from the root 
{\em komen} in this case. As the root {\em komen} specifies that it 
selects such a particle on its {\sc sc}-list, the modal verb 
inherits this specification. The rule  {\tt v\_part\_v} allows us to
combine a verb or verbal complex with a particle. There are two
reasons for not using an analogue of the {\tt vp\_np\_v}-rule in this
case. First, modifiers may not appear in between a particle and the
verbal complex selecting this particle ({\em $^*$ (dat ik) aan om tien
  uur wil komen}). This is accounted for by requiring that the head in
the rule for particles must be {\sc [lex {\em ylex}]} (and 
combinations of a modifier and a verbal head are always {\sc [lex 
  {\em nlex}]}). Second, particles may appear `inside' a verb cluster
({\em (dat ik voor tien uur) zou aan willen komen}, that I would like
to arrive before ten o'clock). This implies that
the result of combining a particle with a verb cluster must be {\sc [lex 
  {\em ylex}]}, instead of {\sc [lex  {\em nlex}]} as specified on
the {\tt vp\_np\_v}-rule.

It should be obvious that these two rules, and the limited form of
argument inheritance we allow (i.e. structure sharing of {\sc
  sc}-lists only, and no concatenation of {\sc sc}-lists), is not
sufficient to account for the full range of verb clustering data in
Dutch. For one thing, the grammar as it stands cannot handle
`inverted' word orders ({\em (dat ik de trein) halen moet}, that I
must catch the train), where the
infinitive precedes the modal verb. It is rather straightforward to
include rules for inverted word orders. A potentially more problematic
omission is the fact that perception verbs ({\em horen, zien}) and
causative {\em laten}, which also introduce verb clusters ({\em (dat
  ik) Rob een kaartje laat kopen}, that I let Rob buy a ticket),
cannot be accounted for. The analysis of this construction in
van Noord and Bouma \shortcite{dvc} is based on the notion `argument inheritance'. This
presupposes the possibility of recursive constraints in syntax (to
concatenate {\sc sc}-lists) as well as rules with an indefinite number of
daughters. Both are excluded within the present formalism. 

\subsubsection{Subordinate clauses}
Subordinate clauses containing a {\sc vp} headed by a finite verb
are of type {\em sbar} (the name {\em sbar} stems
from X-bar grammar, where clauses introduced by a complementizer are 
(barred) projections of {\sc s}).
As finite subordinate clauses are always introduced by a
complementizer, we assume that this complementizer is the head of the
clause and that it subcategorises for a subject {\sc np} and a
(finite) {\sc vp}. The lexical entry for the complementizer {\em dat} (that),
for instance, is:
\enumsentence{
\begin{avm}
{\tt dat} ~~$\mapsto$~~
\[\avmspan{\it comp}\\  sc & \< \@1 \[\avmspan{\it np}\\ case & nom \],
                        \[\avmspan{\it v}\\ vform & fin \\ sc & $\langle\rangle$ \\ 
                                    subj & \<\@1\> \\
                                    sem & \@2  \\
                                    slash & \@3
                                 \] \> \\
                sem & \@2 \\
                mod & $\langle\rangle$ \\
                slash & \@3 \]
\end{avm}
}
\noindent
The complementizer unifies the {\sc np} on its {\sc sc} with the
subject of the {\sc vp}. This implies that the {\sc np} is interpreted
as subject of the {\sc vp}. Furthermore, the complementizer has no
independent semantics, but simply passes on the semantics of the {\sc
  vp}. Since {\em dat} clauses cannot be modifiers, its {\sc mod}
feature is empty. Other complementizers such as {\em omdat} (because) will have
a non-empty value for this attribute to indicate that subordinate
sentences headed by such complementizers can occur as modifier. 

\begin{figure}
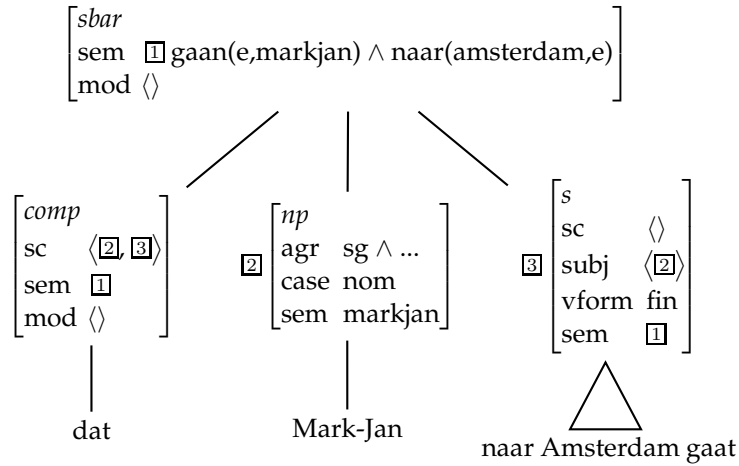

\pstree[levelsep=*1cm,nodesep=3pt]{\Tr[ref=c]{\begin{avm}
    \[\avmspan{\it sbar}\\ 
                   sem & \@1 gaan(e,markjan) $\wedge$
                         naar(amsterdam,e) \\
                   mod & \<\> \]
    \end{avm}
  }}{
\pstree{\Tr[ref=c]{  \begin{avm}
  \[\avmspan{\it comp}\\ sc & \< \@2, \@3 \> \\
                 sem & \@1 \\
                 mod & \<\> \]
\end{avm}
}}{
  \pstree{\Tr[ref=c]{dat}}{}}
\pstree{\Tr[ref=c]{  \begin{avm}
\@2 \[\avmspan{\it np}\\ agr & sg $\wedge$ ... \\
               case & nom \\ 
               sem & markjan  \]
\end{avm}
}}{
  \pstree{\Tr[ref=c]{Mark-Jan}}{}}
\pstree{\Tr[ref=c]{  \begin{avm}
    \@3  \[\avmspan{\it s}\\ sc & $\langle\rangle$ \\
                  subj & \< \@2 \> \\  vform & fin \\
                  sem & \@1 \]
\end{avm}}}{
\pstree[thislevelsep=8pt,edge=none]{\Tfan}{\pstree{\Tr[ref=c]{
      naar Amsterdam gaat}}{}}
           }
}
\centering
\caption{dat Mark-Jan naar Amsterdam gaat (that Mark-Jan is going to Amsterdam)}
\label{sub-ex}
\end{figure}

The 
rule constructing subordinate clauses is defined as follows:\footnote{There is an additional
  rule, for constructing subordinate clauses with a missing
  (`extracted') subject. This
  rule ({\tt sbar2}) could be used in an account of nonlocal
  dependencies which allows for extraction out of subordinate clauses
  as well. }
\begin{numbered}
\begin{avm}
\[\avmspan{\it sbar}\\ slash & \@1 \\
               sem & \@2   \\  mod & \@3   \] ~~$\rightarrow$~~
\[\avmspan{\it comp}\\ sc & \< \@4, \@5 \> \\ slash & \@1 \\ 
               sem & \@2 \\ mod & \@3 \] ~~~
\@4 \[\avmspan{\it np}\\ case & nom \] ~~~
\@5 \[\avmspan{\it v}\\ sc & $\langle\rangle$ \\ subj & \< \@4 \> \\ vform & fin \\ 
            slash & \@1 \\ vslash & $\langle\rangle$ \]
\end{avm}
\end{numbered}
\noindent
A sample derivation is given in figure~\ref{sub-ex}.

\subsubsection{Main clauses}

Main clauses with a finite verb in initial position
(as in yes/no-questions) are of type {\em ques}.
Main clauses in which
the finite verb appears in second position (as in declarative sentences
or {\sc wh}-questions) are of type {\em root}. The attributes
associated with these types are:

\begin{lnumbered}{clause-types}
\begin{avm}
\[\avmspan{\it ques}\\ subj & listof(Sign) \\
    sem & Qlf \\
    slash & listof(Sign) 
    \]~~~~~~~~
\[\avmspan{\it root}\\ sem & Qlf  \]
\end{avm}\end{lnumbered}

Dutch main clauses differ from subordinate clauses in that the finite
verb in main clauses appears in first or second position. There is a
tradition, both in transformational and non-transformational grammar,
to account for this fact by postulating a dependency between the
finite verb and the position where finite verbs occur in subordinate
clauses. The advantage of postulating such a dependency is that the
grammar rules used for subordinate clauses are also applicable in main
clauses. In transformational grammar, a dependency of this type can be
established by means of a head-movement operation which moves the verb
from its final position to a position at the beginning of the
sentence.

Within the framework of {\sc hpsg} \cite{netter-v2,frank} 
we can obtain a similar dependency by postulating a {\em
  verbal trace}, i.e. a verbal sign without phonological content, at
the end of the clause. Using this verbal trace as the head, we can use
the {\sc vp} rules discussed above to build up a {\sc vp} as usual.

\begin{figure}
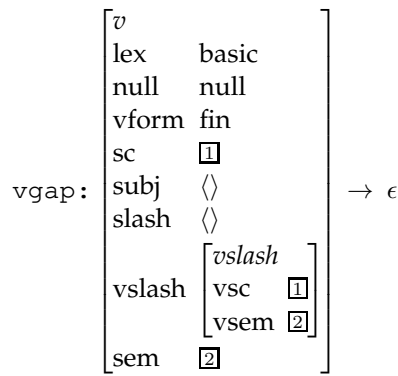

$
\begin{avm}
{\tt vgap:}
\[\avmspan{\it v}\\ lex & basic \\ null & null \\ vform & fin \\
           sc & \@1 \\ subj & $\langle\rangle$ \\ 
            slash & $\langle\rangle$ \\ 
            vslash & \[\avmspan{\it vslash}\\vsc & \@1 \\ vsem & \@2 \] \\
            sem & \@2  \] ~~$\rightarrow$~~$\epsilon$
\end{avm}
$
\centering
\caption{Verbal Gap}
\label{vgap-avm}
\end{figure}

The rule for
introducing such a verbal trace is given in figure~\ref{vgap-avm}. 
Note that the sign for
verbal traces differs from that of an ordinary verb in that its
subcategorisation list in not instantiated, but made reentrant with
{\sc vslash:vsc}. Similarly, the semantics of the verbal gap is
reentrant with {\sc vslash:vsem}. Furthermore, a verbal gap is a basic
(i.e. non-complex) lexical verb, with no phonological content
(i.e. {\sc [null {\em null}]}). We can also safely assume that verbal
traces are finite, as main clauses are always headed by a finite
verb. The value of {\sc subj} is the empty list, as {\sc vp}s headed
by a verbal trace never combine with a subject directly (as will be
shown below). Finally, {\sc slash} also can be assumed to be empty.\footnote{
The lexical rule which moves complements from {\sc sc} to {\sc slash}
does not apply to verbal traces. Instead, it can be applied to the
finite verb which `binds' the trace. Also, if a verbal gap combines
with a complement having a non-empty {\sc slash}, the relevant passing on of the
{\sc slash} value is handled by the finite verb which binds the trace.
This is possible because the {\sc sc}-list of the verbal trace and the binder
will be shared.} 

\begin{figure}
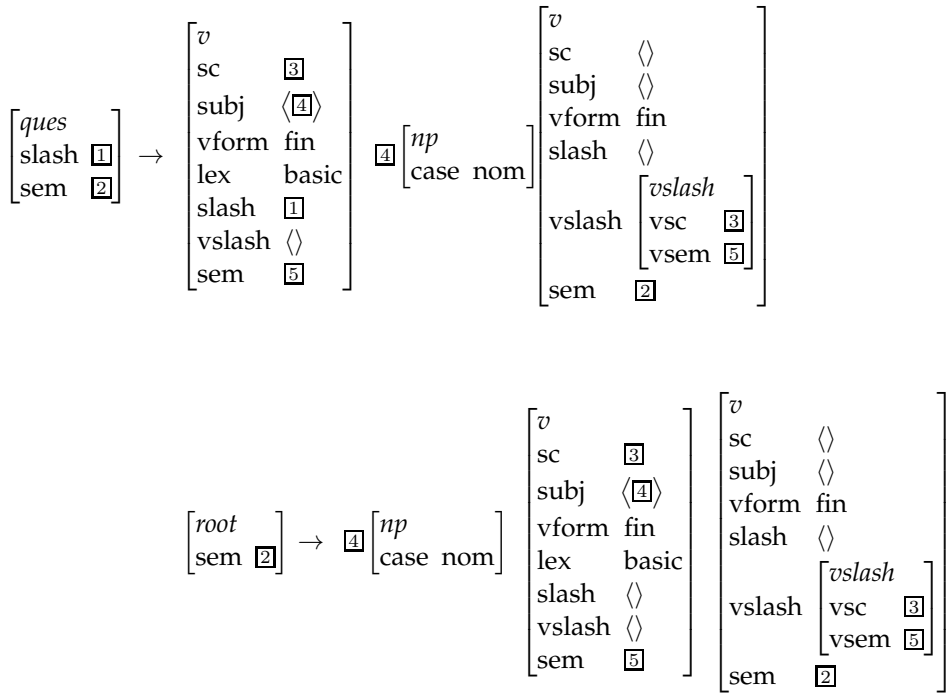

\begin{avm}
\[\avmspan{\it ques}\\ slash & \@1 \\ sem & \@2 \]
~~$\rightarrow$~~
\[\avmspan{\it v}\\ sc & \@3 \\ subj & \< \@4 \> \\ vform & fin \\ lex & basic \\
            slash & \@1 \\ vslash & $\langle\rangle$ \\ sem & \@5 \]~~
\@4 \[\avmspan{\it np}\\ case & nom \]
\[\avmspan{\it v}\\ sc & $\langle\rangle$ \\ subj & $\langle\rangle$ \\ vform & fin \\ 
            slash & $\langle\rangle$ \\ 
            vslash & \[\avmspan{\it vslash}\\ vsc & \@3 \\ vsem & \@5 \]
            \\ sem & \@2 \]
\end{avm}\\[1cm]

\begin{avm}
\[\avmspan{\it root}\\ sem & \@2 \]
~~$\rightarrow$~~
\@4 \[\avmspan{\it np}\\ case & nom \]~~
\[\avmspan{\it v}\\ sc & \@3 \\ subj & \< \@4 \> \\ vform & fin \\ lex & basic \\
            slash & $\langle\rangle$  \\ vslash & $\langle\rangle$ \\ sem & \@5 \]~~
\[\avmspan{\it v}\\ sc & $\langle\rangle$ \\ subj & $\langle\rangle$ \\ vform & fin \\ 
            slash & $\langle\rangle$ \\ 
            vslash & \[\avmspan{\it vslash}\\ vsc & \@3 \\ vsem & \@5 \]
            \\ sem & \@2 \]
\end{avm}
\centering
\caption{Rules for verb-first and subject first main clauses (y/n
  questions and simple declarative sentences)}
\label{mc-avm}
\end{figure}

\begin{figure}
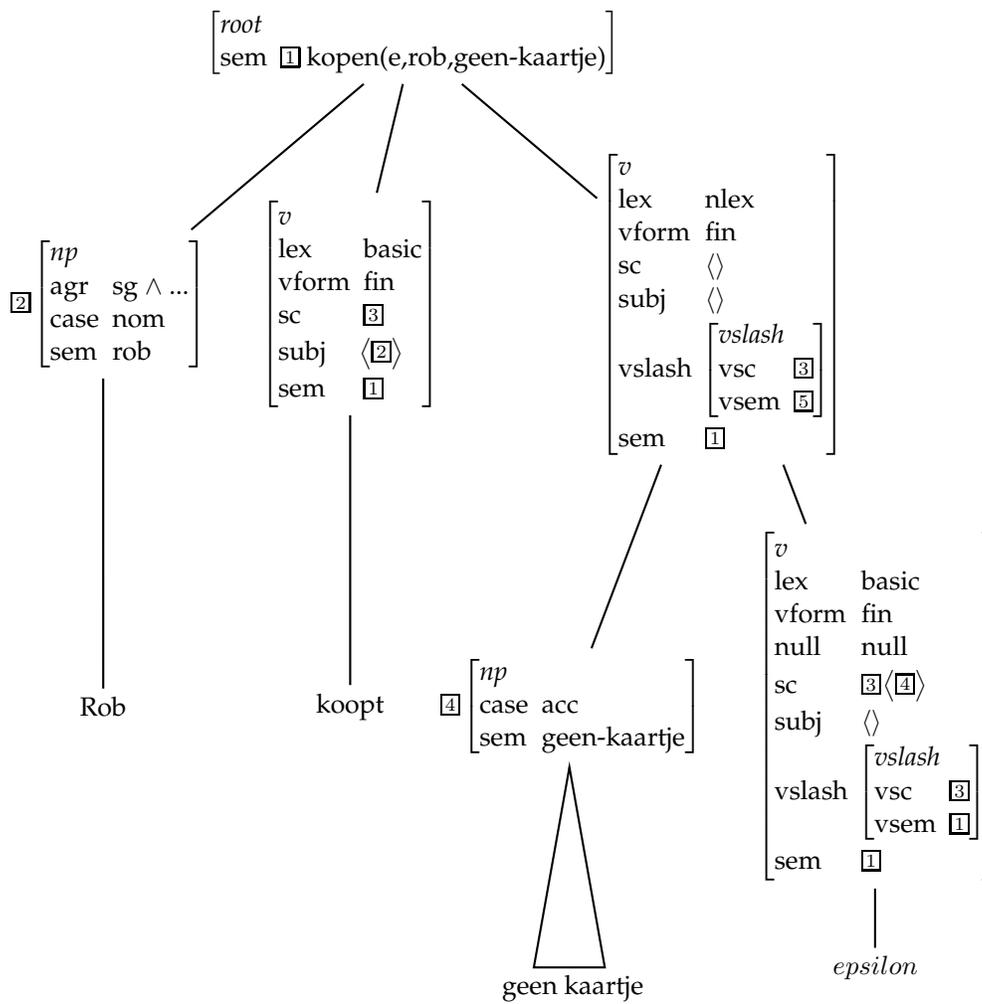

\pstree[levelsep=*1cm,nodesep=3pt]{\Tr[ref=c]{
  \begin{avm}
    \[\avmspan{\it root}\\  sem & \@1 kopen(e,rob,geen-kaartje) \]
    \end{avm}
}}{
\pstree{\Tr[ref=c]{  \begin{avm}
  \@2 \[\avmspan{\it np}\\   agr & sg $\wedge$ ... \\
                 case & nom \\
                 sem & rob  \]
\end{avm}}}{
  \pstree{\Tr[ref=c]{Rob}}{}}
\pstree{\Tr[ref=c]{  \begin{avm}
  \[\avmspan{\it v}\\ lex & basic \\ vform & fin \\ sc & \@3  \\ 
              subj & \<\@2 \> \\ 
                sem & \@1 \]
\end{avm}
}}{
  \pstree{\Tr[ref=c]{koopt}}{}}
\pstree{\Tr[ref=c]{\begin{avm}
\[\avmspan{\it v}\\ lex & nlex \\ vform & fin \\ sc & $\langle\rangle$ \\
              subj & $\langle\rangle$ \\ 
              vslash & \[\avmspan{\it vslash}\\ vsc & \@3  \\ vsem & \@5 \]\\
                sem & \@1  \]
\end{avm}}}{
  \pstree{\Tr[ref=c]{\begin{avm}
\@4 \[\avmspan{\it np}\\ 
               case & acc \\ 
               sem & geen-kaartje  \]
\end{avm}}}{
\pstree[thislevelsep=8pt,edge=none]{\Tfan}{\pstree{\Tr[ref=c]{
      geen kaartje}}{}}
           }
  \pstree{\Tr[ref=c]{  \begin{avm}
      \[\avmspan{\it v}\\lex & basic \\ vform & fin \\ null & null \\ 
                 sc &   \@3 \<\@4 \> \\ subj & $\langle\rangle$ \\ 
                 vslash & \[\avmspan{\it vslash}\\ vsc & \@3 \\ vsem & \@1 \]\\
                sem & \@1 \]
\end{avm}}}{
    \pstree{\Tr[ref=c]{$epsilon$}}{}}}
 }
\centering

\caption{Rob koopt geen kaartje (Rob does not buy a ticket)}
\label{mc-ex}
\end{figure}

There are two rules which combine a finite verb with a {\sc vp}
containing a verbal trace, and which also introduce a subject
(figure~\ref{mc-avm}). Both rules are highly similar (they are
therefore both instances of a {\sc main-clause-struct}). The only
difference is the category of the mother, and the order of the
daughters. The {\em vfirst}-rule introduces
phrases of the type {\em ques}, i.e. instances of verb-first clauses,
in which the subject follows the main verb. The {\em
  subject-first}-rule introduces phrases of type {\em root}, in which
the subject is first, and the main verb follows the subject. The
constraints imply, among others, that the {\sc vp} must
contain a verbal trace, that the {\sc
  sc}-information of the main verb is reentrant with {\sc vslash:vsc}
of the {\sc vp} (and thus, indirectly, with the {\sc sc}-value of the
verbal trace), and that the semantics of main verb is shared with the
value of {\sc vslash:vsem} on the {\sc vp} (and thus, indirectly, with
the semantics of the verbal trace). Note also that the {\sc vp} acts
as semantic head of the construction. This is necessary in order to
ensure that the effect of verbal modifiers within the {\sc vp} is
properly taken into account. An example derivation of a subject first 
main clause is given in figure~\ref{mc-ex}.

\subsubsection{Wh-questions and topicalisation}

In the previous section, we have introduced a rule for verb-initial
and subject-initial main clauses. The first phrase in a main clause
can also be a (non-subject) complement or a modifier. This is
typically the case for (non-subject) {\sc wh}-questions. Sentences with a
`fronted' complement are treated as instances of a non-local
dependency construction (where the dependency is mediated through {\sc
  slash}). In sentences with a fronted modifier, it is assumed
that the first element modifies the remainder of the clause, and thus
a local treatment can be given.

Examples of sentences with a fronted complement are given in (\ref{front-ex}).
\eenumsentence{
\label{front-ex}
\item[a.] Naar welk station wilt u reizen?\\ To which station do you
  want to travel
\item[b.] De laatste trein kunt u nog halen.\\The last train, you can
  still catch
}

These examples are handled by means of a lexical {\sc
  complement-extraction} rule applicable to
verbs, and a syntactic {\sc head-filler}-rule for combining the
fronted element with a {\em ques}-phrase containing a non-empty {\sc
  slash}-value. The {\sc complement-extraction} rule can apply in
two ways: First, it can take a complement  from {\sc sc} and put it on
{\sc slash} (\ref{celr}a). This implies that this complement will not be found
locally, but that it will be unified with an element in `fronted'
position. Second, it can make the {\sc slash} value of a verb
reentrant with the {\sc slash} value of one of its complements (\ref{celr}b). 
This implies that true {\em non-local} dependencies are possible, as
the head of a phrase can pass on information about missing elements
from one of its dependents. If the complement-extraction rule does not
apply, the {\sc slash} value of the verb, as well as the {\sc slash} value of all
its complements, is set to $\langle\rangle$ (the empty list). 
\eenumsentence{
\label{celr}
\item[a.] \begin{avm}
\[\avmspan{\it v}\\ 
   sc   & \< \@1 {\it pp}\> \\ 
   subj & \@2 \\ 
  vform & \@3 \\ 
  $\vdots$ 
\] 
~~$\rightarrow$~~
\[\avmspan{\it v}\\ 
     sc & $\langle\rangle$ \\ 
  slash & \@1 \\ 
   subj & \@2 \\ 
  vform & \@3 \\
  $\vdots$\]
\end{avm} 
\item[b.]
\begin{avm}
\[\avmspan{\it v}\\ 
    sc & \< {\it v}\> \\ 
  subj & \@2 \\ 
 vform & \@3 \\
  $\vdots$ \]
~~$\rightarrow$~~
\[\avmspan{\it v}\\ 
    sc & \<\[\avmspan{\it v}\\ 
             slash & \@1 \] \> \\ 
 slash & \@1 \\ 
  subj & \@2 \\ 
 vform & \@3 \\ 
$\vdots$\]
\end{avm}
}

\begin{figure}
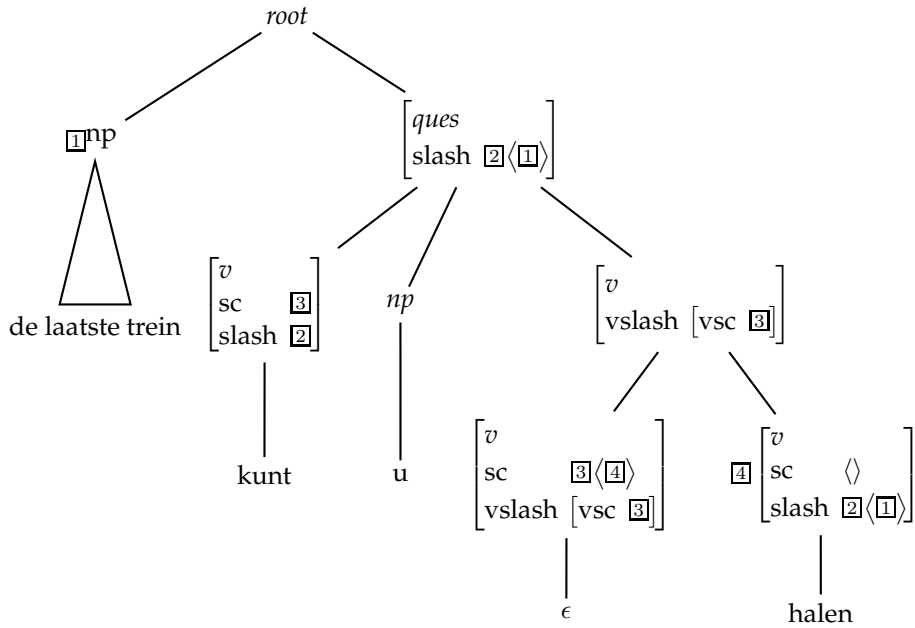


\pstree[levelsep=*1cm,nodesep=3pt]{\Tr[ref=c]{
\em root
}}{ 
 \pstree{\Tr[ref=c]{\begin{avm}   \@1\sort{np}{}  \end{avm}
  }}{
\pstree[thislevelsep=8pt,edge=none]{\Tfan}{\pstree{\Tr[ref=c]{de
      laatste trein}}{}}
}
    \pstree{\Tr[ref=c]{\begin{avm}    \[\avmspan{\it ques}\\ slash & \@2 \< \@1 \> \]
    \end{avm}}}{
       \pstree{\Tr[ref=c]{  \begin{avm}
  \[\avmspan{\it v}\\ sc & \@3  \\ 
              slash & \@2   \]
  \end{avm}
}}{\pstree{\Tr[ref=c]{kunt}}{}}
       \pstree{\Tr[ref=c]{\em np}}{\pstree{\Tr[ref=c]{u}}{}}
       \pstree{\Tr[ref=c]{\begin{avm} \[\avmspan{\it v}\\ vslash & \[ vsc & \@3 \] \]
\end{avm} }}{
         \pstree{\Tr[ref=c]{  \begin{avm}
      \[\avmspan{\it v}\\  sc &   \@3 \< \@4 \> \\ vslash & \[ vsc & \@3 \] \]
\end{avm}}}{\pstree{\Tr[ref=c]{$\epsilon$}}{}}
         \pstree{\Tr[ref=c]{ \begin{avm}
\@4 \[\avmspan{\it v}\\ sc & $\langle\rangle$ \\ slash & \@2 \< \@1 \>  \]
\end{avm}}}{\pstree{\Tr[ref=c]{halen}}{}}}}  
  }
\centering

\caption{De laatste trein kunt u halen}
\label{wh-deriv}
\end{figure}

An example of a derivation involving {\sc slash} is given in
figure~\ref{wh-deriv}. The {\sc complement extraction} rule has
applied to {\em halen} (to catch) to produce a verbal sign with an empty {\sc
  sc}-list and an {\em np} on {\sc slash}. A verbal trace contains a
reentrancy between its {\sc sc}-list and its {\sc vslash:vsc}-list. When
the verbal trace combines with {\em halen},
the information that {\em halen} has an {\em np} on {\sc slash} will
therefore also be instantiated on {\sc vslash:vsc}. This information
is passed up to the resulting verb phrase. The {\em complement
  extraction} rule also applies to the finite verb {\em kunt} (can), but in
this case it establishes a reentrancy between the {\sc slash} value of
the verb on the {\sc sc}-list of {\em kunt} and the {\sc slash}-value
of {\em kunt} itself. The {\sc vfirst} rule unifies the {\sc sc}-list
of {\em kunt} with the {\sc vslash:vsc}-list of the verb phrase {\em
  halen $\epsilon$}, and thus, {\sc slash} (of the verb on {\sc sc}
of {\em kunt}, and thus on {\em kunt} itself) is instantiated as
$\langle np \rangle$. This information is passed on to the resulting
{\em ques} phrase, which can then be combined with the initial {\em
  np} using the {\sc topicalisation} rule in \ref{topic-rule}.

\begin{lnumbered}{topic-rule}
\begin{avm}
\[\avmspan{\it root}\\sem & \@1 \] ~~$\rightarrow$~~
\@2 ~~
\[\avmspan{\it ques}\\slash & \@2 \\ sem & \@1 \]
\end{avm}
\end{lnumbered}

It should be noted that our account of non-local dependencies differs
from earlier {\em slash}-based accounts, such as those in
Gazdar et al.\ \shortcite{gpsg} and Pollard and Sag \shortcite{hpsg2} 
in that it does not make use of a
{\sc foot feature} principle. Instead, we adopt the approach of
Sag \shortcite{Sag-Relatives}, who imposes the canonical constraint that
the {\sc slash}-value of a head is the set-union of the {\sc
  slash}-values of its daughters. An {\sc extraction} lexical rule can
be used to remove an element from {\sc sc} ({\sc comps}) and to add
this element to the set of elements on {\sc slash}. In our
implementation, we have made several simplifying assumptions. First,
{\sc slash} is not a set, but a list. Second, this list can contain at
most one element. This assumption (which has the effect of restricting
the number of `missing' elements from a phrase to at most one) is too 
restrictive for a highly limited number of cases in English, but
appears to be valid for Dutch. Third, instead of imposing a general
constraint that {\sc slash} must be the concatenation on the {\sc
  slash} values of all elements on {\sc sc}, we allow the {\sc
  complement extraction} rule to unify the value of {\sc slash} with
one specific element on {\sc sc}. We have to make this assumption, as
the more general alternative requires the use of {\em delayed
  evaluation}, something which we wish to avoid in this
grammar, or {\em difference lists}. While the latter alternative is
possible within the present formalism, it also introduces a number of
complications which are avoided in the present implementation. The
fourth and final simplification is that {\sc complement extraction}
and {\sc slash} feature passing is only possible for verbs. This is
certainly too restrictive, as extraction out of subordinate clauses of
type {\em sbar} ({\em welke trein zegt Gertjan dat Rob gemist
  heeft?}, which train does Gertjan say that Rob has missed) and out
of {\em pp}s ({\em Waar gaat deze trein naar toe?}, Where does this
train go to), and a number of other types of phrase is possible as well.

Sentences where the first phrase is a modifier are dealt with without
appealing to {\sc slash}. Instead, it is assumed that in sentences
such as (\ref{mod-ex}), the fronted elements modify the following {\em
  ques} phrase. This requires an additional (mod-topic) rule, given
in~(\ref{root-mod}). 
\eenumsentence{
\label{mod-ex}
\item[a.] Hoe laat gaat de volgende trein naar Zwolle?\\When does the
  next train to Zwolle leave?
\item[b.] Woensdag moet ik om tien uur in Zwolle zijn.\\Wednesday, I
  must be in Zwolle by ten o'clock.
}

\begin{lnumbered}{root-mod}
\begin{avm}
\[\avmspan{\it root}\\sem & \@1 \] ~~$\rightarrow$~~
\[\avmspan{\it modifier}\\ sem & \@1 \\ mod & \< \@2 \> \] ~~
\[\avmspan{\it ques}\\slash & $\langle\rangle$ \\ sem & \@2 \]
\end{avm}
\end{lnumbered}
\noindent 
Of course, this account rests on the assumption
that modifiers of embedded verbs or phrases cannot be fronted, an
assumption which is almost certainly false in general (see
Hukari and Levine \shortcite{hukari-adjuncts}, for instance), but which appears to be rather
unproblematic for present purposes.

\subsubsection{Special grammar rules}
\label{special}

The domain which has been selected for \ovis\ (information dialogues
concerning public transportation) and the fact that \ovis\ deals with
spoken language, imply that it is crucial that a number of grammatical
phenomena are described in a robust manner. In particular, temporal
expressions, locative expressions (names of cities and stations), and
a number of typical spoken language constructions, such as greetings,
occur frequently in such dialogues. 

The grammar rules and lexical entries for these phenomena make use of
the \ovistwo\ grammar formalism, but are
not organised according to the linguistic principles discussed above.
This is true not only for the syntax, but also for semantics. The
reason for dealing with these phenomena by means of a set of more or
less {\em ad hoc} rules and lexical entries is that the constructions
discussed below are often extremely idiosyncratic. At the same time,
describing the regularities that can be observed does not seem to
require the overhead of the grammar architecture we assume for the
rest of the grammar. The most economical and robust solution seemed
therefore to encapsulate the grammar for these constructions in
relatively independent grammar modules.

\subsection{Semantics}
\label{semantics}

The output of the grammatical analysis is a semantic, linguistically
motivated and domain-independent, representation of the utterance, in
the form of a Quasi Logical Form ({\sc qlf}). The {\sc qlf} formalism
was developed in the framework of the {\em Core Language Engine} ({\sc
  cle}, \cite{cle-book,AlshawiCrouch}). Since then, the formalism was used and
further developed in projects such as the {\em Spoken Language
  Translator} \cite{slt94}, {\em Clare} \cite{clare}, in the {\em
  Fracas}-project \cite{fracas8} and in {\em Trace \& Unification
  Grammar} \cite{block94}. In {\sc ovis} the {\sc qlf} is translated
into a domain-specific {\em update} expression, which is passed on to
the pragmatic interpretation module and dialogue manager for further
processing. The dialogue manager maintains an information state to
keep track of the information provided by the user. An {\em update}
expression is an instruction for updating the information state
\cite{tst24}.  Below, we motivate our choice for {\sc qlf}s as
semantic representation language and we discuss how these {\sc qlf}s
are translated into updates.

\subsubsection{The semantic representation language}

Predicate logic, (sometimes extended with for example {\em generalised
  quantifiers} or {\em discourse markers}), is often used to represent
the meaning of sentences.  Due to its long tradition in describing
semantics of natural languages it is now a well established and well
understood technique. The main advantage of artificial languages like
predicate logic is that they are unambiguous. An ambiguous natural
language utterance will therefore correspond to more than one
expression in predicate logic, one for each reading of the utterance.
The disadvantage of this approach is that for very ambiguous inputs,
expensive computations must be carried out to compute all readings.
The alternative adopted in formalisms based on the idea of {\em
  monotonic semantic interpretation} (\cite{fracas8}, see also
\cite{nerbonne-semantics2} and \cite{logic_and_lexicon}) is to
represent ambiguity by means of under-specification and to postpone the
computation of individual readings as long as possible.

Representing ambiguity by under-specification, and postponing the 
computation of individual readings, has at least two computational
advantages. First, parsing can benefit significantly from the fact that
ambiguities which are only semantic (i.e. do not have a syntactic 
counterpart) are represented by a single derivation. Second, ambiguity
resolution can often proceed without enumerating all possible readings
of an input separately. A striking example of the latter situation is
the translation of {\sc qlf}'s that are ambiguous with respect to 
quantifier-scope into a domain-specific meaning representation as  
it is used by the dialogue manager of the {\sc ovis}-system. 
The utterance in (\ref{neg}a), for instance, gives rise to a single 
{\sc qlf} (\ref{neg}b), which could be resolved (ignoring the existential 
quantification over events and the fact that it is a question) to either 
(\ref{neg}c) or (\ref{neg}d). The domain-specific reading of (\ref{neg}a) 
(which corresponds to (\ref{neg}c)) is computed on the basis of 
(\ref{neg}b) directly, and thus never needs to consider
the two different readings of this {\sc qlf}.

\eenumsentence{
\label{neg}
\item[a.] \begin{flushleft}Gaat er niet een latere (trein)?\\
                           Is there not a later train?\\[10pt]
          \end{flushleft}
\item[b.]
\begin{avm}
\[ pred & not\\
   args & \< \[ pred & leave\\
                args & \< \[ index & $e_1$
                          \], 
                          \[ index & \@3 \\
                             res & $\lambda$\@4.later\_train(\@4)\\
                             q & exist
                          \]
                       \>
             \] 
          \>
\]
\end{avm}\\
\item[c.] $not(\exists x$ (later\_train($x$) $\wedge$ leave($e_1,x$)))\\
\item[d.] $\exists x$ (later\_train($x$) $\wedge$ not(leave($e_1,x$)))
}

\subsubsection{Quasi logical form}

\begin{figure}[h,t,b]
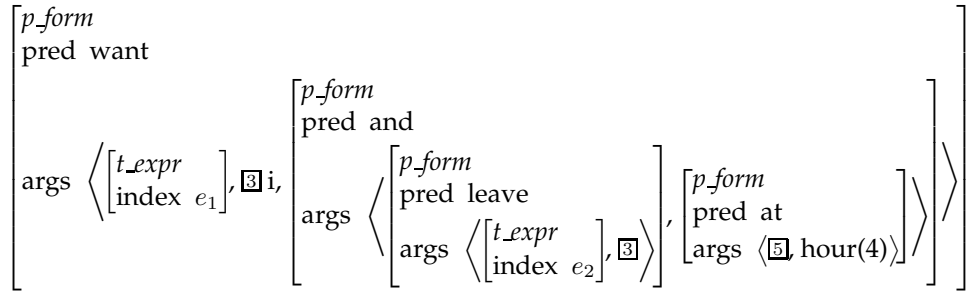

\begin{avm}
\[ \avmspan{\em p\_form} \\
   pred & want\\
   args & \< \[ \avmspan{\em t\_expr} \\
                index & $e_1$
             \],
             \@3 i,
             \[ \avmspan{\em p\_form} \\
                pred & and \\
                args & \< \[ \avmspan{\em p\_form} \\
                             pred & leave\\
                             args & \< \[ \avmspan{\em t\_expr} \\
                                          index & $e_2$
                                       \], \@3
                                    \>
                          \],
                          \[ \avmspan{\em p\_form}\\
                             pred & at\\
                             args & \< \@5, hour(4) \>
                          \]
                       \>
             \]
            
          \>
\]
\end{avm}
\centering
\caption{{\sc qlf} for 'Ik wil om ongeveer vier
  uur vertrekken' (I want to leave at about four o'clock)} 
\label{fig1}
\end{figure}

In figure \ref{fig1} we give a {\sc qlf} as it is produced by the
{\sc ovis}-grammar. It is a typed feature-structure, whose
main components are predicative forms ({\em p\_form}), representing
relations (which may also be higher order, such as {\em not} and 
{\em and}), and terms. Generalised quantifiers are  represented
by term expressions ({\em t\_expr}).  The example in (\ref{fig1}) 
contains two generalised quantifiers, corresponding to the (existentially
quantified) event-variables introduced by the two verbal predicates 
\cite{davidson67}. Note that these quantifiers appear as arguments of
the predicates, and thus are unscoped with respect to each other.

Our implementation of {\sc qlf} in the {\sc ovis} grammar follows
roughly the presentation in \cite{fracas8}, although some of the
apparatus supplied for contextual resolution in that work has been
omitted.  As the {\sc ovis}-grammar uses typed feature-structures, 
{\sc qlf}'s are represented as feature-structures below. 

A {\sc qlf} is either a {\em qlf-term} or a {\em qlf-formula}. \
A {\it qlf-term} is one of the following:

\begin{itemize}
\item a term index,\footnote{In the original formalism {\em indices}
    and {\em variables} are distinguished. An index uniquely
    identifies a {\tt term} expression. At this moment indices and
    variables have the same function in our implementation. We
    may need to distinguish between them later.}
\item a constant term,
\item an term-expression of type {\em t\_expr} and containing the features
  {\sc index, restr} and {\sc quant}\footnote{
In chapter 5 of \cite{fracas8} term expressions also
      contain a slot {\sc cat} for specifying information about the lexical form
      and syntactic/semantic type of an expression (e.g. quantifier,
      pronoun, etc.) and a slot {\sc ref} for specifying the
      (contextual) referent of an expression. We do use {\sc cat}, but
      have omitted it from the presentation below. We currently do not
      use {\sc ref}.}
(see (\ref{fig1})), where {\sc index} is a variable, 
  {\sc restr} is an expression of predicate logic (possibly
  with lambda-abstraction) and {\sc quant} is a generalised quantifier.
\end{itemize}
 
A {\sc qlf} formula is one of the following:\footnote{In chapter 5 of
  \cite{fracas8} two more formula constructs are introduced. These are
  not used in the current implementation.}

\begin{figure}
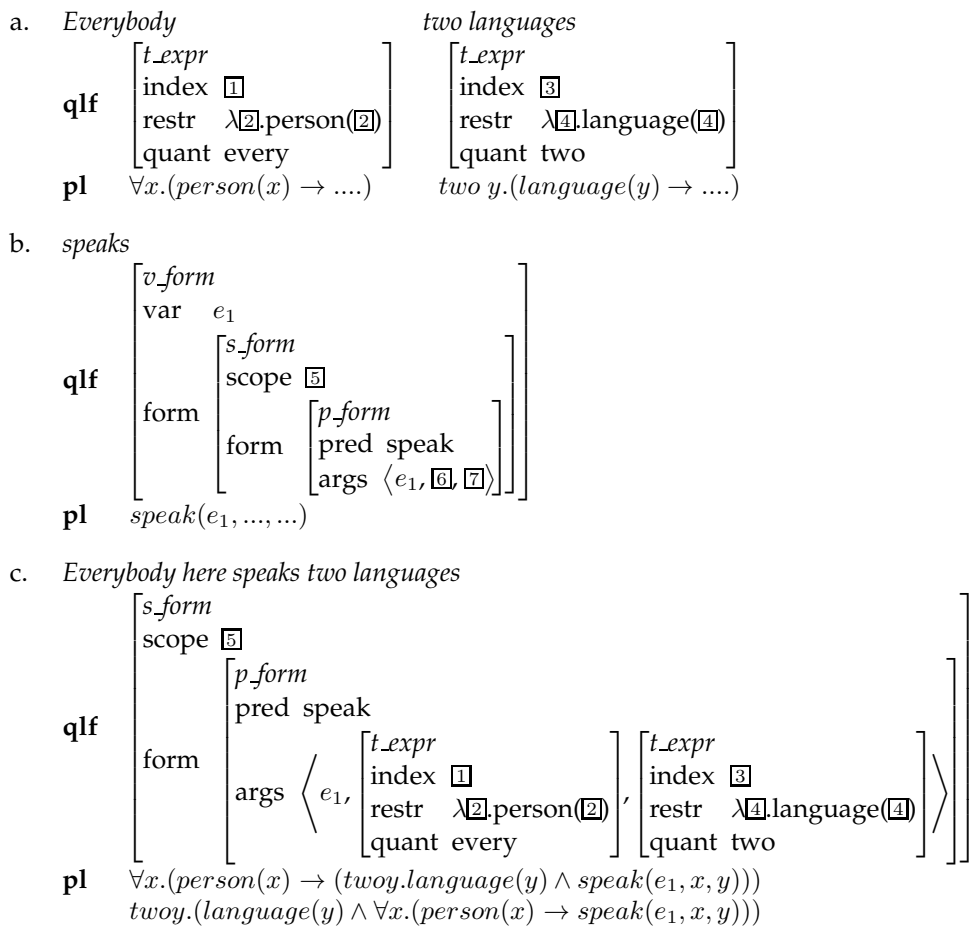


\begin{tabular}{lll}

a. &  \multicolumn{2}{l}{\em Everybody \hspace{31mm}  two languages}\\
& {\sc \bf qlf} &  
\begin{avm}
\[
\avmspan{\em t\_expr}\\
index & \@1 \\
restr & $\lambda$\@2.person(\@2)\\
quant & every 
\]
\end{avm}
\hspace{5mm}
\begin{avm}
\[
\avmspan{\em t\_expr}\\
index & \@3 \\
restr & $\lambda$\@4.language(\@4)\\
quant & two
\]
\end{avm}\\
& {\sc \bf pl} &  $\forall x.(person(x) \rightarrow ....)$ \hspace{7mm}
 ${\sc two}~y.(language(y) \rightarrow ....)$ \\[10pt]

b. &
\multicolumn{2}{l}{\em speaks}\\
& {\sc \bf qlf} & 
\begin{avm}
\[ \avmspan{\em v\_form}\\
   var & $e_1$\\
   form & \[ \avmspan{\em s\_form}\\
             scope & \@5 \\
             form  & \[ \avmspan{\em p\_form}\\
                        pred & speak \\
                        args & \< $e_1$, \@6, \@7 \> 
                     \]
          \]
\]
\end{avm} \\[10pt]
& {\sc \bf pl} & $speak(e_1,...,...) $\\[10pt]
c. &
\multicolumn{2}{l}{\em Everybody here speaks two languages}\\
& {\sc \bf qlf} & 
\begin{avm}
\[ \avmspan{\em s\_form}\\
   scope & \@5 \\
   form  & \[ \avmspan{\em p\_form}\\
              pred & speak \\
              args & \< $e_1$, \[ \avmspan{\em t\_expr}\\
                           index & \@1 \\
                           restr & $\lambda$\@2.person(\@2)\\
                           quant & every
                         \],   
                         \[ \avmspan{\em t\_expr}\\
                            index & \@3 \\
                            restr & $\lambda$\@4.language(\@4)\\
                            quant & two 
                         \]\> 
                     
          \]
\] 
\end{avm}\\[10pt]
& {\sc \bf pl} 
& $\forall x.(person(x) \rightarrow({\sc two} y.language(y) \wedge  speak(e_1,x,y)))$\\
& & $ {\sc two} y.(language(y) \wedge \forall x.(person(x) \rightarrow
 speak(e_1,x,y)))$
\end{tabular}

\caption{The relation between an expression in {\sc qlf} and a
  fomula of predicate logic}
\label{qlfpl}
\end{figure}

\begin{itemize}
\item a predicate-argument formula of type {\em p\_form}, and with
  features {\sc pred} and {\sc args} (see (\ref{fig1})). Predicates may
  be higher order, arguments may be formulas or terms,
\item a formula of type {\em v\_form} with features {\sc var} and {\sc
    form} representing a formula with lambda-abstraction
  (see(\ref{qlfpl}b)). This is an 
  auxiliary level of representation, introduced to facilitate the
  interaction between grammar-rules and lexical entries, 
\item a formula of type {\em s\_form} (see(\ref{qlfpl}b)), with
  features {\sc scope} and {\sc form}. The value of {\sc scope}  is
  either a variable or a list of indices indicating the relative scope
  of term expressions (generalised quantifiers) (see (\ref{qlfpl}c)).
\end{itemize}

The definitions can best be illustrated with a simple example in which
we compare a {\sc qlf} expression with its corresponding formula in
predicate logic. In figure \ref{qlfpl} the sentence {\em Everybody
  speaks two languages} is given both a translation in {\sc qlf} and in predicate
logic. In the {\sc qlf}-translation of the full sentence the scope
order (\begin{avm}\@5\end{avm})
of the two quantifiers is left unspecified. Resolving scope order
amounts to instantiating \begin{avm}\@5\end{avm} to
\begin{avm}[\@1,\@3]\end{avm} (for everybody there are two languages
that s/he speaks) or to \begin{avm}[\@3,\@1]\end{avm} (there are two
languages that everybody speaks).

\subsubsection{Construction of {\sc qlf}'s}

During grammatical analysis {\sc qlf}s are constructed
compositionally (see also \cite{AlshawiCrouch}). 
In {\em head complement structures} the head daughter
is the syntactic as well as the semantic head of the structure. This
means that the semantic content of the complement constituents is
combined with the semantic content of the head. The value of the {\sc
  sem} feature of the head is passed up to the mother (see figure
\ref{principles}).

In {\em head modifier structures} the modifier is the semantic
head. The semantics of the syntactic head of the structure is plugged
into the {\sc mod} feature of the modifier. Below we will show
how the semantics of the modifier is combined with the semantics of
the constituent it modifies. The value of the {\sc sem} feature of the
modifier is passed up to the mother.

\begin{figure}[t,b,p]
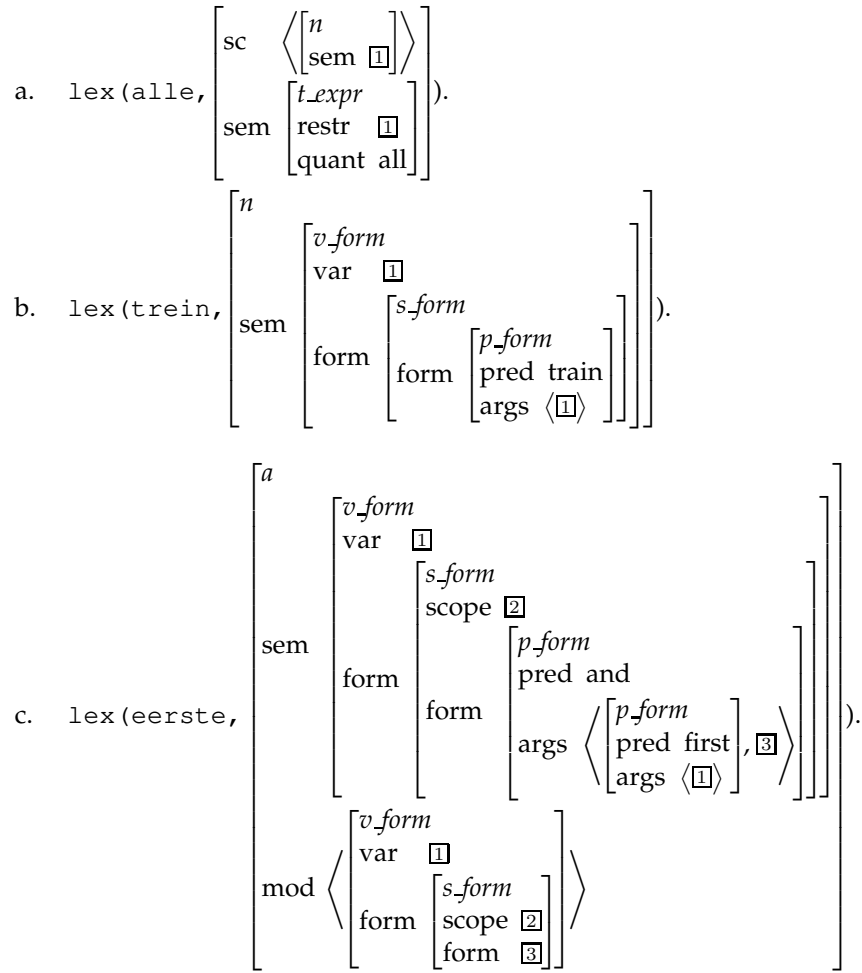


\begin{tabular}{ll}
a. & 
\begin{avm}
\mbox{\tt lex(alle,}\[
sc & \< \[\avmspan{\it n}\\ sem & \@{1}
        \]
     \>\\
sem & \[\avmspan{\it t\_expr}\\
         restr & \@{1}\\
         quant & all
      \]
\] ).
\end{avm}\\[30pt]
b. & 
\begin{avm}
\mbox{\tt lex(trein,}\[\avmspan{\it n}\\
   sem & \[\avmspan{\it v\_form}\\
            var & \@1\\
            form & \[\avmspan{\it s\_form}\\
                      form & \[\avmspan{\it p\_form}\\
                                 pred & train \\
                                 args & \<\@1 \>
                             \]
                   \]
         \]
\]\mbox{).}
\end{avm}\\[50pt]
c. & 
\begin{avm}
\mbox{\tt lex(eerste,}
\[\avmspan{\it a}\\
 sem & \[\avmspan{\it v\_form}\\
            var & \@1\\
            form & \[\avmspan{\it s\_form}\\
                      scope & \@2\\
                      form & \[\avmspan{\it p\_form}\\
                                 pred & and\\
                                 args & \< \[\avmspan{\it p\_form}\\
                                                 pred & first \\
                                                 args & \< \@1 \>
                                              \], \@3
                                           \>
                             \]
                   \]
         \]\\
   mod & \<\[\avmspan{\it v\_form}\\
              var & \@1\\
              form & \[\avmspan{\it s\_form}\\ 
                         scope & \@2\\
                         form & \@3
                     \]
           \]
         \>
\]\mbox{).}
\end{avm}
\end{tabular}
\caption{\label{nsems}Examples showing the semantics of determiners,
  nouns, and adjectives. }
\end{figure}

We now discuss the semantics of various linguistic categories. 
{\em Determiners} subcategorise for a noun (see Figure
\ref{nsems}(a.)). The semantics of the noun is unified with 
restriction of the determiner. Nouns introduce a {\em v\_form} (fig. \ref{nsems}(b.))
Note that it is also assumed that quantifiers may scope at this point. 
{\em Adjectives} are modifiers (fig.\ref{nsems}(c.)). They operate on structures whose
semantic content is of type {\em v\_form}. The lambda variables of the
two formulas are unified and the semantic content of the structure is
the conjunction of the logical formula of the adjective and the
logical formula of the structure it modifies.

\begin{figure}
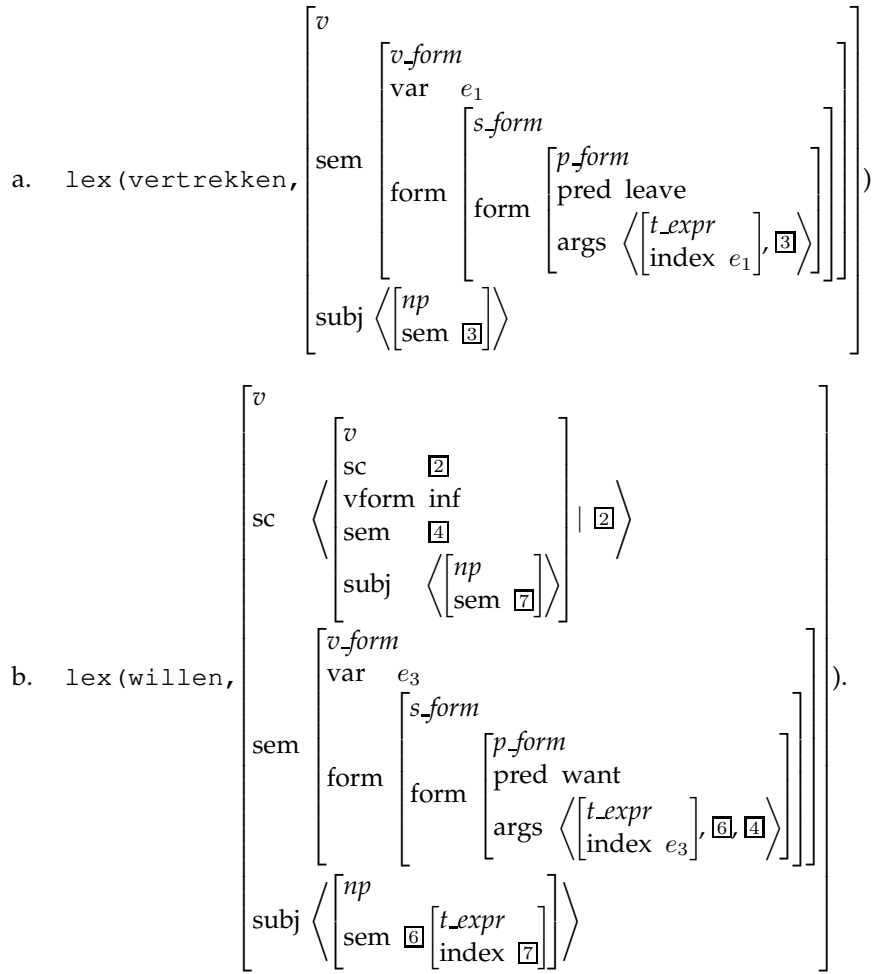

\begin{tabular}{ll}
a. & 
\begin{avm}
\mbox{\tt lex(vertrekken,}\[\avmspan{\it v}\\
   sem & \[\avmspan{\it v\_form}\\
            var & $e_1$\\
            form & \[\avmspan{\it s\_form}\\
                      form & \[\avmspan{\it p\_form}\\
                                pred & leave \\
                                args & \< 
      \[\avmspan{\it t\_expr}\\
         index & $e_1$
      \], \@3
                                          \>
                             \]
                   \]
         \]\\
   subj & \< \[\avmspan{\it np}\\
                sem & \@3
             \]
          \>
\]\mbox{).}
\end{avm}
\\[70pt]
b. & 
\begin{avm}
\mbox{\tt lex(willen,}\[\avmspan{\it v}\\
sc & \< \[\avmspan{\it v}\\
           sc & \@2\\
           vform & inf\\
           sem & \@4 \\
           subj & \< \[ {\it np}\\
           sem & \@7
                  \] \>
        \] \| \@2
     \> \\
sem & \[\avmspan{\it v\_form}\\
         var & $e_3$\\
         form & \[\avmspan{\it s\_form}\\
                   form & \[\avmspan{\it p\_form}\\
                             pred & want \\
                             args & \< \[\avmspan{\it t\_expr}\\
                                             index & $e_3$
                                          \],
                                          \@6, \@4
                                       \>
                          \]
                \]
      \]\\
subj & \< \[\avmspan{\it np}\\
           sem & \@6 \[ \avmspan{\em t\_expr} \\ index & \@7 \]
        \]
     \>
\]).
\end{avm}
\end{tabular}
\caption{Verbal semantics.}
\label{vsems}
\end{figure}

The semantics of {\em verbs} corresponds with a {\em v\_form} (see
fig. \ref{vsems}a). The value
of {\sc var} is reentrant with the {\sc index} of the event introduced
by the verb. The semantics of the subject is unified with the second 
element of the argument list of the verb. Intransitive verbs have
two semantic arguments, corresponding to the event and subject,
respectively. Transitive verbs have three arguments, where the third
argument is unified with the semantics of the single element on {\sc
  sc}. 

{\em Modal verb} are {\em subject-control} verbs. This means that the
subject of the {\sc vp}-complement is controlled by the subject of the
modal verb. Semantically, this means that the {\sc index} of the
subject must be unified with the semantics of the subject of the {\sc
  vp}-complement. Note also that we assume that assume that the {\sc
  sc}-list of a modal verb may contain complements introduced by the 
{\sc vp}-complement (as explained in section \ref{vps}). These are not
relevant for the semantics of the modal verb.

\begin{figure}
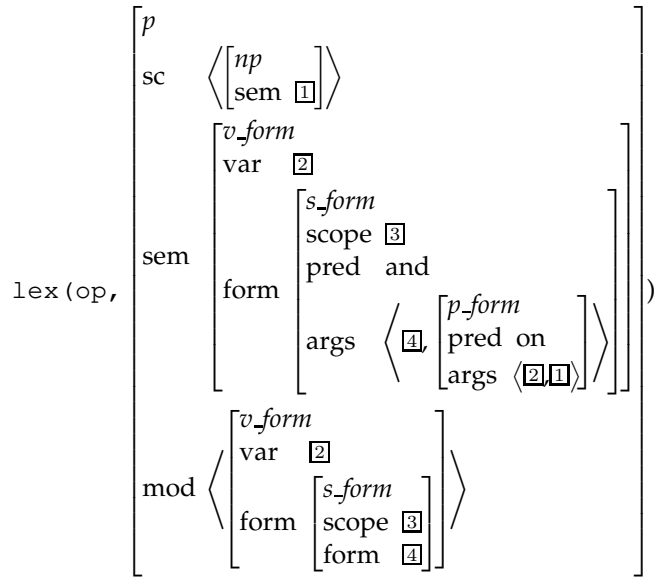

\begin{avm}
\mbox{\tt lex(op,}
\[\avmspan{\it p}\\
sc & \< \[\avmspan{\it np}\\ sem & \@1\]\>\\
sem & \[\avmspan{\it v\_form}\\
           var & \@2\\
           form & \[\avmspan{\it s\_form}\\
                      scope & \@3\\
                      pred & and\\
                      args & \< \@4, \[\avmspan{\it p\_form}\\
                                           pred & on\\
                                           args & \<\@2,\@1\>
                                        \]
                                \>
                  \] 
      \]\\
mod & \< \[\avmspan{\it v\_form}\\
            var & \@2\\
            form& \[\avmspan{\it s\_form}\\
                     scope & \@3\\
                     form & \@4
                  \]
         \]
      \>
\]).
\end{avm}
\centering
\caption{\label{psems} Adverbial semantics for prepositions. }
\end{figure}

The semantics of {\em adverbial phrases} resembles 
that of {\em adjectives}. In figure (\ref{psems}) the semantics
of prepositions heading a {\sc pp} which acts as a verbal modifier is
given. {\sc Pp}-modifiers introduce a conjunction, with the verbal
semantics as first argument, and the prepositional semantics as
second. The {\sc index} of the {\sc vp} is the first argument of the 
predicate introduced by the preposition, the semantics of the {\sc
  np}-object of the preposition corresponds to the second argument.

In Dutch, temporal {\sc np}'s can act as verbal modifiers: 
\eenumsentence{
\item[a] \begin{flushleft}
         Ik wil {\em zondag} vertrekken\\
         I want to leave on Sunday
         \end{flushleft}
\item[b] \begin{flushleft}
         Ik wil {\em drie januari} naar Amsterdam\\
         I want to go to Amsterdam on the third of January
         \end{flushleft}
\item[c] \begin{flushleft}
         Ik wil er uiterlijk {\em drie uur} zijn\\
         I want to arrive at the latest at three o'clock
         \end{flushleft}
}

\begin{figure}
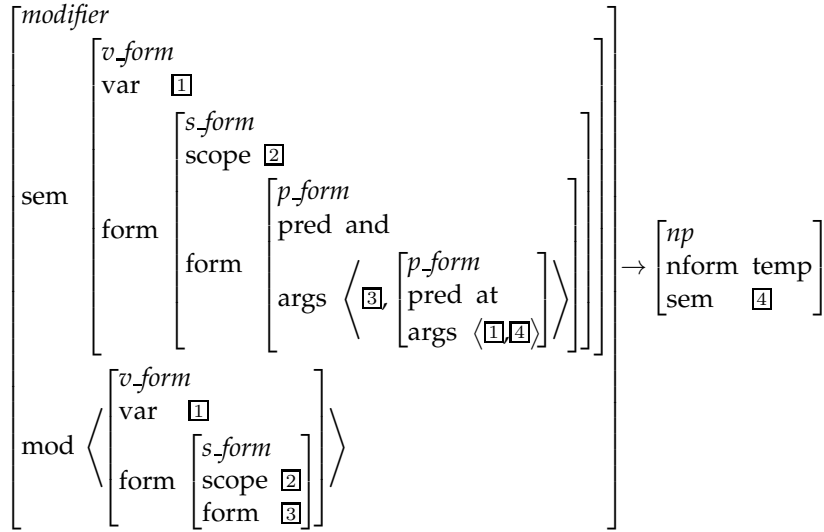

\begin{avm}
\[\avmspan{\it modifier}\\
sem & \[\avmspan{\it v\_form}\\
         var & \@1\\
         form & \[\avmspan{\it s\_form}\\
                   scope & \@2\\
                   form & \[\avmspan{\it p\_form}\\
                             pred & and\\
                             args & \< \@3,
                          \[\avmspan{\it p\_form}\\
                             pred & at\\
                             args & \<\@1,\@4\>
                          \]
                                       \>
                          \]
                \]
      \]\\
mod & \< \[\avmspan{\it v\_form}\\
            var & \@1\\
            form & \[\avmspan{\it s\_form}\\
                      scope & \@2\\
                      form & \@3
                   \]
         \]
      \>
\] 
$\rightarrow$ 
\[\avmspan{\it np}\\
   nform & temp\\
   sem   & \@4
\]
\end{avm}
\centering
\caption{Rule {\tt mod\_np} to treat temporal noun phrases as modifiers. }
\label{np2modsem}
\end{figure}

As {\sc np}s normally do not have a modifier semantics, there is a
unary rule that transforms temporal {\sc np}s into modifiers
(figure~\ref{np2modsem}). The structure that is modified is specified
in the {\sc mod} feature.  The semantic content of the modifier is
constructed as if it was a {\sc pp} with {\sc p\_form} {\em om} (at).
The semantic content of the (temporal) {\sc np} daughter is plugged
into the second position of
the argument list of the preposition.\\

\subsection{Constructing updates from {\sc qlf}s}
\label{updates}

The dialogue manager keeps track
of the information provided by the user by maintaining an {\em
  information state\/} or {\em form} \cite{tst24}. This form is a hierarchical
structure, with slots and values for the origin and destination of a
connection, for the time at which the user wants to arrive or depart,
etc. An example is given in (\ref{update-ex}a).
Each user utterance leads to an {\em update} of the information
state. An update is an instruction for updating the information in an
information state. Updating can mean that new information is added or
that given information is confirmed, retracted or corrected. For
example, given the information state in (\ref{update-ex}a), the update
in (\ref{update-ex}b) (which might be the translation of 
{\em No, I do not want to travel to Leiden but to
  Abcoude!}) leads to the information state in (\ref{update-ex}c). The
\# -operator in (\ref{update-ex}b) indicates that the information within its
scope (indicated by square brackets) is to be retracted, and the 
'!'-operator indicates a correction. 

\eenumsentence{
\label{update-ex}
\item[a.]
\begin{avm}
\[ travel & \[ origin & \[ place & \[ town & groningen \] \\
                           moment & \[ at & \[ time & \[ clock\_hour & 3 \]
                                            \] \]
                         \]\\
               destination & \[ place & \[ town & leiden \] \]
             \]
\]
\end{avm}
\item[b.] travel.destination.
                  ([\# place.town.leiden];
                   [! place.town.abcoude])
\item[c.]
\begin{avm}
\[ travel & \[ origin & \[ place & \[ town & groningen \] \\
                           moment & \[ at & \[ time & \[ clock\_hour & 3 \]
                                            \] \]
                         \]\\
               destination & \[ place & \[ town & abcoude \] \]
             \]
\]
\end{avm}
}

The result of parsing is a {\sc qlf}, a linguistically motivated and
domain-independent representation of the meaning of a sentence. The 
translation of a {\sc qlf} into a domain-specific {\em update} is done
by applying translation-rules to the individual parts of a {\sc
  qlf}. These translation rules may be context-sensitive. In
particular, some parts of the {\sc qlf} provide the context which
determines how other parts are to be translated. For example, the {\sc
  qlf} in (\ref{fig1a}) (corresponding to the phrase {\em leave at
  four o'clock} contains two {\em p\_forms}, one for the predicate {\em
  leave} and one for {\em four o'clock}. The second gives rise to an
update expression {\tt moment.at.time.clock\_hour.4}. The first
provides the contextual information that the moment referred to is a
departure-time. The translation can therefore be extended to 
{\tt origin.moment.at.time.clock\_hour.4}. There is no linguistic
information which indicates that a special update-operator has to be
used. In such cases, it is assumed that the information is new, and
thus the assert-operator ('=') can be used, giving rise to the
translation for the full phrase:
{\tt origin.moment.at.[= time.clock\_hour.4]}.

\enumsentence{
\label{fig1a}
\begin{avm}
             \[ \avmspan{\em p\_form} \\
                pred & and \\
                args & \< \[ \avmspan{\em p\_form} \\
                             pred & leave\\
                             args & \< \[ \avmspan{\em t\_expr} \\
                                          index & $e_1$
                                       \], \@3
                                    \>
                          \],
                          \[ \avmspan{\em p\_form}\\
                             pred & at\\
                             args & \< $e_1$, hour(4) \>
                          \]
                       \>
             \]
\end{avm}
}

Contextual translation is a powerful technique. For instance, the
utterance {\em Groningen Amsterdam} gives rise to a conjunctive {\sc
  qlf}, containing two term expressions for locations. Translating each of the
conjuncts individually would make it impossible to decide whether an
origin or destination location is being specified. By translating the
conjunction in one step (and assuming that the order of conjuncts
corresponds to the order in the utterance), we can resolve the first 
locative to origin and the second to destination. 
As another example, the adverb {\em graag} is ignored in the
translation from {\sc qlf} to update if it occurs as part of a full
sentence ({\em ik wil graag naar Amsterdam}, `I would like to go to
Amsterdam'), but is translated as `yes' (i.e. a confirmation of
information provided by the system) if it occurs in isolation. 
Such a translation is motivated by dialogues of the following type:

\enumsentence{
\begin{tabular}[t]{ll}
[system:] & Dus U wilt van Amsterdam naar Groningen reizen?\\
          & So you want to travel from Amsterdam to Groningen?\\
        
[user:] & Graag. \\
        & Please.
\end{tabular}
}

Similarly, the translation of the negations {\em nee} (no) and {\em
  niet} (not) depends on context. If the two occur in isolation, they
indicate a denial of information provided by the system. However, if 
{\em nee} is followed by another
phrase, say a locative, it signals a correction (\ref{neeniet}a), whereas
if {\em niet} is followed by another phrase, it signals a denial 
(\ref{neeniet}b).

\eenumsentence{
\label{neeniet}
\item[a.] 'Nee, naar Assen' ({\em No, to Assen})\\
{\tt destination.[!place.assen]}
\item[b.] 'Niet naar Assen' ({\em Not to Assen})\\
{\tt destination.[\#place.assen]}
}

It should be noted that the translation of {\sc qlf}'s to updates uses
primarily the information provided by {\sc np}'s, {\sc pp}'s 
and adverbs. Verbs typically provide the context for translating other
parts of the {\sc qlf}. Also, as quantification plays no role in
updates, the scope of generalised quantifiers can be largely
ignored. Thus, we are able to translate {\sc qlf}'s into
domain-specific meanings without resolving quantifier scope.

\section{Robust parsing of word-graphs}
\label{parser}

\subsection{Word-graphs}
\begin{sidewaysfigure}
\includegraphics[scale=0.35]{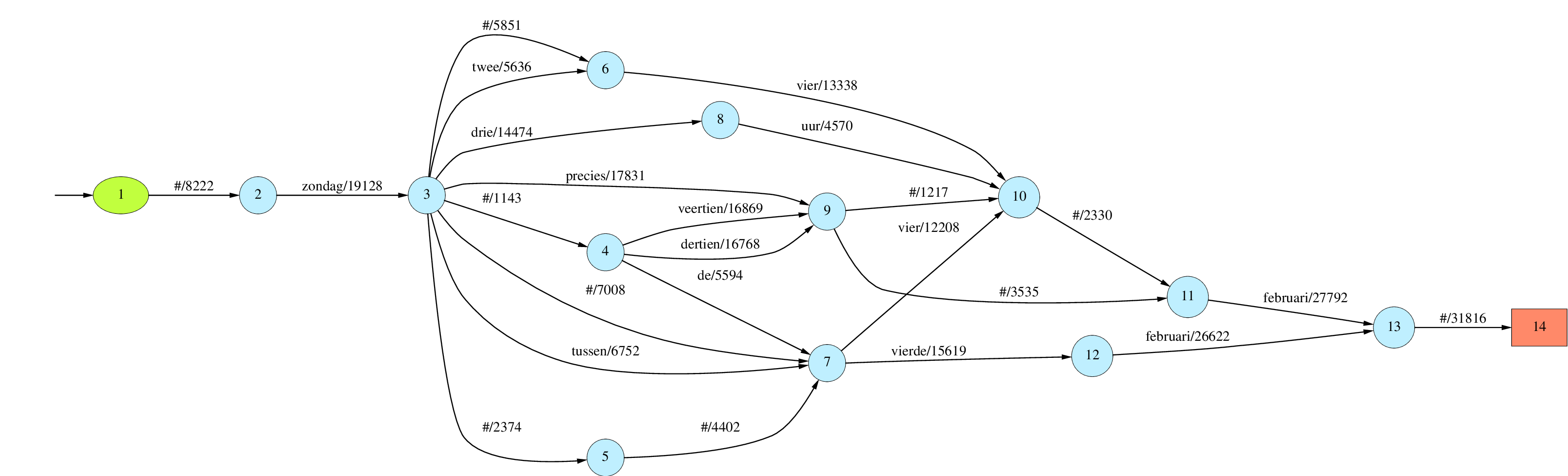}
\includegraphics[scale=0.35]{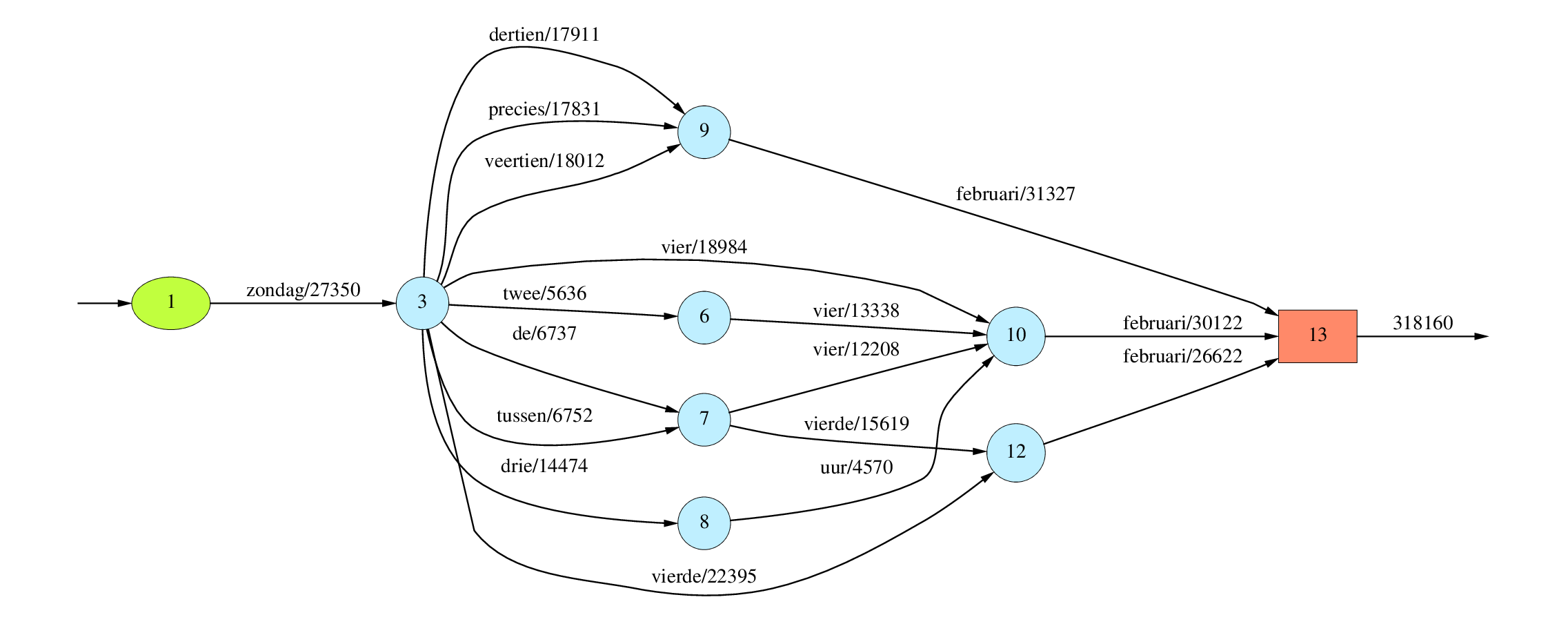}
\centering
\caption{\label{wgkps}Word-graph and normalized word-graph
  for the utterance {\em Zondag vier
    februari} (Sunday Februari fourth). The special label \#
  in the first graph indicates a pause transition. These transitions
  are eliminated in the second graph. }
\end{sidewaysfigure}

The input to the NLP module consists of word-graphs produced by the
speech recogniser \cite{oerder-ney}. A word-graph is a compact
representation for all sequences of words that the speech recogniser
hypothesises for a spoken utterance. The states of the graph represent
points in time, and a transition between two states represents a word that
may have been uttered between the corresponding points in time. Each
transition is associated with an {\em acoustic score} representing a measure
of confidence that the word perceived there  was
actually uttered. These scores are negative logarithms of
probabilities and therefore require addition as opposed to
multiplication when two scores are combined.  An example of a typical
word-graph is given as the first graph in figure~\ref{wgkps}.

At an early stage, the word-graph is normalised to eliminate the {\em
  pause transitions}. Such transitions represent periods of time for
which the speech recogniser hypothesises that no words are uttered.
After this optimisation, the word-graph contains exactly one start
state and one or more final states, associated with a score,
representing a measure of confidence that the utterance ends at that
point. The word-graphs in figure~\ref{wgkps} provide an example.

From now on, we will assume word-graphs
are normalised in this sense.  Below, we refer to transitions in the
word-graph using the notation $\mbox{\it trans\/}(v_i,v_j,w,a)$ for a
transition from state $v_i$ to $v_j$ with symbol $w$ and acoustic
score $a$. Let $\mbox{\it final\/}(v_i,a)$ refer to a final state $v_i$ with
acoustic score $a$.

\subsection{Parsing word-graphs}
The normalized word-graph is parsed by an appropriate parser. Parsing
algorithms for strings can be generalized to parse such word-graphs
(for some examples cf. van Noord \shortcite{acl95}).  
In the ideal case, the parser
will find a path in the word-graph that can be assigned an analysis
according to the grammar, such that the path covers the complete time
span of the utterance, i.e.\ the path leads from the start state to a
final state. The analysis gives rise to an update of the dialogue
state, which is then passed on to the dialogue manager.

However, often no such paths can be found in the word-graph,
due to:
\begin{itemize}
\item errors made by the speech recognizer,
\item linguistic constructions not covered in the grammar, and
\item irregularities in the spoken utterance.
\end{itemize}

Even if no full analysis of the word-graph is possible, it is usually 
the case that useful information can be extracted from the word-graph.
Consider for example the utterance: 
\enumsentence{\label{vanvan}
  Ik wil van van Assen naar Amsterdam\\
  I want from from Assen to Amsterdam 
} 
The grammar will not assign an analysis to this utterance due to the
repeated preposition. However, it would be useful if the parser would
discover the prepositional phrases {\em van Assen} and {\em naar
  Amsterdam} since in that case the important information contained in
the utterance can still be recovered. Thus, in cases where no full
analysis is possible we would like to fall back on an approach
reminiscent of concept spotting. The following proposal implements
this idea.

Firstly, the grammar is defined in
such a way that each {\em maximal projection} such as {\sc s},
{\sc np}, {\sc pp}, etc., can be analysed as a top category. This
is well-motivated because utterances very often consist of a single {\sc
  np} or {\sc pp} (section~\ref{robgr}). 

Often, the task of the parser is to discover all instances of the top
category from the start state of the word-graph to a final state. 
But in our case, we require that the parser discovers all
instances of the top category {\em anywhere in the word-graph},
i.e. for all partial paths in the word-graph. This has the desired
effect for example~(\ref{vanvan}): both {\sc pp}s will be found by the
parser. 

Thus we require that the parser finds all major categories anywhere in
the word-graph.  If a bottom-up chart parser is used, then we might
use the inactive chart items for this purpose. However, since we do
not want to be forced to a particular parsing strategy, we have chosen
to adopt a different approach. In section~\ref{robpa} we show that
in a logic programming setting the use of underspecification of the
state names associated with the top-most goal obtains the desired
effect, without loss of efficiency. 

Therefore, after the parser has finished, we have a word-graph
annotated with a number of instances of top categories. For each of
these categories we are interested in the word-graph state where this
category starts ($v_i$), the word-graph state where this category ends
($v_j$), the sequence of symbols associated with this category ($x$),
the accumulated acoustic score ($a$), and the qlf ($q$). Let
$\mbox{\it parsed\/}(v_i,v_j,x,a,q)$ refer to such categories.

We are interested in paths from the start state to the final state
consisting of a number of categories and transitions in the word-graph
(the latter are called {\em skips}). The problem consists in finding
the optimal path, according to a number of criteria. This problem is
formalized by defining the annotated word-graph as a directed acyclic
graph (section~\ref{robdef}).  The vertices of this graph are the
states of the word-graph; the edges are the transitions of the
word-graph and the categories found by the parser. 

The criteria which are used to favor some paths over other paths are
expressed as a weight function on the edges of the graph.  The
criteria we might take into account are discussed in
section~\ref{robwe}. For instance, a typical criterion will favor
paths consisting of a small number of categories, and a small number
of skips. The case in which the parser found a full analysis from
the start state of the word-graph to a final state 
then reduces to a special case: the analysis solely consisting of that
category will be favored over sequences of partial analyses.

Obviously, it is not a good idea to generate all possible sequences of
categories and skips, and then to select the best path from this set:
in typical word-graphs there are simply too many different paths.  If
a certain uniformity requirement on weights is met, however, then
efficient graph search algorithms are applicable. The particular
algorithm implemented in \ovistwo, namely a variant of the {\sc
  dag-shortest-path} algorithm \cite{algorithms}  is
discussed in section~\ref{robal}.

The criteria used to determine the best path may also include Ngram
statistics. It turns out that in those cases some complications arise
in the definition of the annotated word-graph. This is explained in
section~\ref{robco}.

In a previous implementation \cite{madrid} we used a version of
Dijkstra's algorithm. A comparison is presented in
section~\ref{robalt}. Finally, section~\ref{bfm} discusses methods in
which the parser is applied only to a single path of the word-graph. 

\subsection{Grammar}
\label{robgr}
We require that grammatical analysis finds all maximal
projections anywhere in the input word-graph. This implies that the
top category of the grammar should be defined in such a way that it
derives each of these maximal projections. For this reason, the
grammar contains the declaration:
\begin{numbered}\begin{verbatim}
top_category(X) :- X => start.
\end{verbatim}\end{numbered}

Furthermore, there are unary rules rewriting this start category into
each of the relevant maximal projections. One such rule is:
\begin{numbered}\begin{verbatim}
rule(start_np,Start,[Np]) :-
    Start => start, Np => np,
    Start:sem <=> Np:sem.
\end{verbatim}\end{numbered}
Similar rules are defined for {\tt pp, sbar, root, advp}, etc.

\subsection{Parser}
\label{robpa}
Five different parsing algorithms were implemented and compared (a
bottom-up Earley parser, an inactive chart parser, an LR parser, a
left-corner parser and a head-corner parser).  The most efficient
parser (both in terms of CPU-time and memory usage) for this
application turned out to be a head-corner parser implemented with
goal-weakening and selective memoization.  The head-corner parser is
presented in detail in van Noord \shortcite{cl97}.

In order to apply this (or any of the other) parser(s) for robust
processing, we use underspecification of the state names for the
input parse goal in order to parse the start category {\em anywhere
  in the word-graph}.  Normally the parser will be called using a
goal such as the following:
\begin{numbered}\begin{verbatim}
?- parse(start(Sem),q0,q16).
\end{verbatim}\end{numbered}
indicating that we want to find a path from state {\tt q0} to {\tt
  q16} which can be analysed as a category {\tt start(Sem)} (a
sentence with a semantic representation that is yet to be discovered).
If we want to recognize top categories at all positions in the input,
then we can simply generalize the parse goal to:
\begin{numbered}\begin{verbatim}
?- parse(start(Sem),_,_).
\end{verbatim}\end{numbered}

Now it may seem that such an underspecified goal will dramatically
slow down the parser, but this turns out to be a false expectation, at
least for the head-corner and left-corner parsers. In fact we
have experienced no such decrease in efficiency. This can only be
understood in the light of the use of memoization: even though we now
have a much more general goal, the number of different goals that we
need to solve is much smaller.

\subsection{Annotated word-graph}
\label{robdef}
An annotated word-graph is a word-graph annotated with the results of
the parser. Such an annotated word-graph is defined with respect to
an input word-graph (given by the functions $\mbox{\it trans\/}$ and
$\mbox{\it final\/}$) and with respect to the results of parsing (given
by the function $\mbox{\it parsed\/}$).  

The {\em annotated word-graph} is a directed acyclic graph $(V,E)$ where
\begin{itemize}
\item $V$ is the set of vertices consisting of the states of the
word-graph $v_0\dots v_n$, and a new vertex
$v_{n+1}$. $v_0$ is the start state. $v_{n+1}$ is the final state.
\item $E$ is the set of edges consisting of:
\begin{enumerate}
\item {\em skip} edges. For all $\mbox{\it trans\/}(v_i,v_j,w,a)$
  there are edges $(v_i,v_j,w,a,\epsilon)$. 
\item {\em category} edges.
  For all $\mbox{\it parsed\/}(v_i,v_j,x,a,q)$
  there are edges $(v_i,v_j,x,a,q)$. 
\item {\em stopping} edges.
  For all $\mbox{final}(v_i,a)$ there are
  edges $(v_i,v_{n+1},\epsilon,a,\epsilon)$. 
\end{enumerate}
\end{itemize}

\subsection{Weights}
\label{robwe}
The weights that are associated with the edges of the graph can be
sensitive to the following factors.

\begin{itemize}
\item Acoustic score. Obviously, the acoustic score 
  present in the word-graph is an important factor. The
  acoustic scores are derived from probabilities by taking the negative
  logarithm. For this reason we aim to minimize this score. If edges
  are combined, then we have to sum the corresponding acoustic scores.
\item Number of `skips'. We want to minimize the number of  skips,
  in order to obtain a preference for the maximal projections found by
  the parser. Each time we select a skip edge, the number of skips is
  increased by 1.
\item Number of maximal projections. We want to minimize the number of
  such maximal projections, in order to obtain a preference for more
  extended linguistic analyses over a series of smaller ones. Each
  time we select a category edge, this number is
  increased by 1.
\item Quality of the {\sc qlf} in relation to the context. We are
  experimenting with evaluating the quality of a given {\sc qlf} in
  relation to the dialogue context, in particular the question
  previously asked by the system \cite{koeling-tblisi}.
\item Ngram statistics. We have experimented with bigrams and
  trigrams. Ngram scores are expressed as negative logarithms of
  probabilities. This implies that combining Ngram scores requires
  addition, and that lower scores imply higher probability.
\end{itemize}

The only requirement we make to ensure that efficient graph searching
algorithms are applicable is that weights are {\em uniform}. This
means that a weight for an edge leaving a vertex $v_i$ is independent
of how state $v_i$ was reached.  

In order to be able to compute with such multidimensional weights, we
express weights as tuples $\langle c_1, \dots, c_k\rangle$. For each
cost component $c_i$ we specify an initial weight, and we need to specify for
each edge the weight of each cost component. To specify how weights are
updated if a path is extended, we use the function $\mbox{\it uw\/}$
that maps a 
pair of a multidimensional weight and an edge a to multidimensional
weight.  \footnote{We do not define a weight function on edges, but
  we specify how weights are updated if a path is extended, for
  generality. This approach allows e.g.\ for the possibility that
  different cost components employ different operations for combining weights.
  For example, some cost components may use addition (e.g. for weights which
  are expressed as negative logarithms derived from probabilities),
  whereas other cost components may require multiplication (e.g. for
  probabilities).}  Moreover, we need to define an ordering on such
tuples. In order to experiment with different implementations of this
idea we refer to such a collection of specifications as a {\em
  method}. Summarizing, such a weight method is a triple
$W=\langle\mbox{\it ini\/},\mbox{\it uw\/},\prec\rangle$ where
\begin{enumerate}
\item $\mbox{\it ini\/}$ is the initial weight;
\item $\mbox{\it uw\/}$ is the update weight function;
\item $\prec$ is an ordering on weights
\end{enumerate}

\paragraph{Example: speech method.}
As a trivial example of such a method, consider the problem of finding
the best path through the word-graph ignoring all aspects but the
acoustic scores present in the word-graph.  In order to implement a
method $W^{\mbox{speech}}$ to solve this problem, we define weights
using a unary tuple $\langle c\rangle$. The initial weight is
$\mbox{\it ini\/}=\langle 0\rangle$ and $\mbox{\it uw\/}$ is defined as
follows:
\begin{equation}
\mbox{\it uw\/}(\langle c\rangle,(v_i,v_j,w,a,q)) =
\left\{ \begin{array}{cl} 
\langle c + a\rangle & \mbox{for skip edges}\\
\langle\infty\rangle & \mbox{for category edges}\\
\langle c+a\rangle   & \mbox{for stopping edges}\\
\end{array}\right.
\end{equation}

Note that we specify an infinite weight for category edges because we
want to ignore such edges for this simple method (i.e. we are simply
ignoring the results of the parser).  We define an ordering
$\prec$ on such tuples, simply by stating that
$\langle w\rangle \prec \langle w'\rangle$ iff $w < w'$.

\paragraph{Example: nlp\_speech method.}
A more interesting example is provided by the following method which
not only takes into account acoustic scores, but also the number of
skip edges and category edges. Weights are expressed as 
$\langle c_1, c_2, c_3\rangle$, where $c_1$ is the number of skips,
$c_2$ is the number of categories, and $c_3$ is the acoustic score.

We define $\mbox{\it ini\/} = \langle 0, 0, 0\rangle$ and
$\mbox{\it uw\/}$ is defined as follows.
\begin{equation}
\mbox{\it uw\/}(\langle c_1, c_2, c_3\rangle,(v_i,v_j,w,a,q)):
\left\{ \begin{array}{cl}
\langle c_1+1,c_2, c_3 + a\rangle  &   \mbox{for skip edges}\\
\langle c_1, c_2+1, c_3 + a\rangle &   \mbox{for category edges}\\
\langle c_1, c_2, c_3+a\rangle     &   \mbox{for stopping edges}\\
\end{array}\right.
\end{equation}

Finally, we define the ordering on such tuples: 
\begin{equation}
\langle c_1, c_2, c_3\rangle \prec \langle c'_1, c'_2, c'_3\rangle \mbox{ iff}:
\left\{ \begin{array}{l} 
    c_1 < c'_1 \mbox{ or}\\
    c_1 = c'_1 \mbox{ and } c_2 < c'_2 \mbox{ or}\\
    c_1 = c'_1 \mbox{ and } c_2 = c'_2 \mbox{ and } c_3 < c'_3\\
\end{array}\right. 
\end{equation}

\subsection{Search algorithm}
\label{robal}
The robustness component can be characterised as a search in the
annotated word-graph. The goal of the search is the best path from
$v_0$ to $v_{n+1}$.  This search reduces to a well-known graph search
problem, namely the problem of finding the shortest path in a directed
acyclic graph with {\em uniform} weights.

We use a variant of the {\sc dag-shortest-path} algorithm
\cite{algorithms}.  This algorithm finds shortest paths
for uniformly weighted directed acyclic graphs. The first step of the
algorithm is a topological sort of the vertices of the graph. It turns
out that the state names of the word-graph that we obtain from the
speech recogniser are already topologically sorted: state names are
integers, and edges always connect to larger integers.  The second
step of the algorithm maintains an array $A$ which records for each
state $v_k$ the weight associated with the best path known from $v_0$
to $v_k$. A similar array, $P$, is used to represent for each state
the history of this best path, as a sequence of {\sc qlf}s (since
that is what we want to obtain eventually).

The first step of the algorithm initialises these arrays such that
for each state $v_i (i\neq 0) A[v_i] = \infty$, and $P[v_i] =
\mbox{NIL}$. For $v_0$ we specify $A[v_0] = \mbox{\it ini\/}$ and
$P[v_0]=\epsilon$. After this initialisation phase the algorithm
treats each edge of the graph in topologically sorted order of the
source vertex, as follows:
\begin{numbered}
\normalsize
\begin{flushleft}
\tt foreach state $v_i$ (in topologically sorted order)\\
\hspace{4em}do\\
\hspace{4em}foreach edge $(v_i,v_j,w,a,q)$\\
\hspace{8em}    do\\
\hspace{8em}    relax $(v_i,v_j,w,a,q)$\\
\end{flushleft}
\end{numbered}

The function {\tt relax} is defined on edges and updates the arrays if
a better path to a vertex has been found:
\begin{numbered}
\normalsize
\begin{flushleft}
\tt function relax $(v_i,v_j,w,a,q)$\\
\hspace{4em}    if $\mbox{\it uw\/}(A[v_i],(v_i,v_j,w,a,q)) \prec A[v_j]$\\
\hspace{4em}    then $A[v_j] =:  \mbox{\it uw\/}(A[v_i],(v_i,v_j,w,a,q))$\\
\hspace{6.5em}         $P[v_j] =: P[v_i].q$\\
\end{flushleft}
\end{numbered}
When the algorithm finishes, $P[v_{n+1}]$ constitutes the sequence of
{\sc qlf}s associated with the best path found. The weight of this
path is given by $A[v_{n+1}]$. This algorithm is efficient. Its
running time is O($V+E$), where $V$ is the number of vertices and $E$
is the number of edges. Therefore, it can be expected that this part
of processing should not decrease parsing efficiency too much, since
the number of edges is O($V^2$).\footnote{This compares well with the
  O($V^3$) complexity which can be obtained for most parsers. } For a
more detailed account of the correctness and complexity of this
algorithm, see Cormen, Leiserson, and Rivest \shortcite{algorithms}. \footnote{Note that the algorithm is
  different from the Viterbi algorithm. The latter algorithm finds the
  best path through a possibly cyclic weighted graph, {\em for
  a given sequence of observed outputs}. In the current application we
  require an algorithm to find the best path in an {\em acyclic}
  weighted graph (without an additional observed output sequence).}

A simple generalisation of the algorithm has been implemented in order
to obtain the N best solutions. In this case we maintain in the
algorithm for each vertex the N best paths found so far. Such a
generalisation increases the worst-case complexity by only a constant
factor, and is very useful for development work.

\subsection{Complications for Ngrams}
\label{robco}

In this section we want to extend the {\em nlp\_speech} method to take
into account Ngram probabilities. Obviously, we can extend the weight
tuple with a further argument which expresses the accumulated weight
according to the Ngram probabilities. However, a potential problem
arises. If we extend a given path by using the transition labeled
$w$, then we want to take into account the probability of
reading this word $w$ given the previous $N-1$ words. However note that in
the definition of the annotated word-graph given above these words are not
readily available. Even if we make sure that each path is associated
with the last words seen so far, we must be careful that  weights
remain uniform. 

The solution to this problem is to alter the definition of the graph,
in such a way that the relevant $N-1$ words are part of the vertex.
If we want to take into account Ngram probabilities ($N=2,3, \dots$),
then the vertices of the graph will be tuples $(v,w_1\dots w_{N-1})$
where $v$ is a state of the word-graph as before, and $w_1\dots
w_{N-1}$ are the previous $N-1$ words.  For example, in the case of
bigrams ($N=2$), vertices are pairs consisting of a word-graph state
and a word (the previous word).  A number of special symbols $y_{N-1}
\dots y_1$ is introduced as beginning-of-sentence markers. The start
vertex is now $(v_0,y_{N-1}\dots y_1)$. The notation $x:k$ is used to
refer to the last $k$ elements of $x$.

The {\em annotated word-graph for Ngrams} is a weighted graph
$(V,E)$ and some fixed $N$, such that:
\begin{itemize}
\item $V$ is a set of pairs $(v,w_1\dots w_{N-1})$ where $v$ is a
  word-graph state and $w_i$ are labels in the word-graph. 
The start vertex is $(v_0,y_{N-1}\dots y_1)$; the final vertex is 
$(v_{n+1},\epsilon)$.
\item $E$ is the set of edges consisting of:
\begin{enumerate}
\item {\em skip} edges. For all $\mbox{\it trans\/}(v_i,v_j,w,a)$, 
and all vertices $V_i=(v_i,x)$ and
  $V_j=(v_j,xw:N-1)$, there are edges $(V_i,V_j,w,a,\epsilon)$. 
\item {\em category} edges.  For all
$\mbox{\it parsed\/}(v_i,v_j,x_2,a,q)$, and for
all vertices $V_i=(v_i,x_1)$  and $V_j=(v_j,x_1x_2:N-1)$, 
there are edges $(V_i,V_j,x_2,a,q)$.
\item {\em stopping} edges.  For all $\mbox{final}(v_i,a)$ and for
  all vertices $V_i=(v_i,x)$ there are edges
  $(V_i,(v_{n+1},\epsilon),\epsilon,a,\epsilon)$.
\end{enumerate}
\end{itemize}

\paragraph{Example: nlp\_speech\_trigram method.}
The start state of the graph search now is the vertex $(v_0,
y_2y_1)$. Weights are expressed as 4-tuples by extending the
triples of the {\em nlp\_speech} method with a fourth component
expressing trigram weights. These trigram weights are expressed using
negative logarithms of (estimates of) probabilities. Let $tri$ be the
function which returns for a given sequence of three words this
number. Moreover, the definition of this function is extended for
longer sequences of words in the obvious way by defining 
$tri(w_0w_1w_2x) = tri(w_0w_1w_2) + tri(w_1w_2x)$. 

The initial weight is defined as $\mbox{\it ini\/}=\langle 0,0,0,0\rangle$. 
Weights are updated as follows:

\begin{numbered}\normalsize
\begin{flushleft}
$\mbox{\it uw\/}(\langle c_1, c_2, c_3, c_4\rangle,((v_i,w_0w_1), 
(v_j,y),x,a,q)) =$ \\ 
\hspace{5em}$\left\{ \begin{array}{cl} 
\langle c_1+1,c_2,c_3 + a,c_4 + tri(w_0w_1x)\rangle &  
                  \mbox{for skip edges}\\ 
\langle c_1,c_2+1,c_3 + a,c_4 + tri(w_0w_1x)\rangle &  
                  \mbox{for category edges}\\  
\langle c_1,c_2,c_3+a, c_4\rangle                   &  
                  \mbox{for stopping edges}\\ 
\end{array}\right. $  
\end{flushleft} 
\end{numbered} 

Finally, we define an ordering on such tuples. The function $\mbox{\it
  total}$ is defined on tuples as follows. Here $k_{\mbox{\scriptsize
    nlp}}$ and $k_{\mbox{\scriptsize wg}}$ are constants.

\begin{equation}
\mbox{\it total\/}(\langle c_1, c_2, c_3, c_4\rangle) =
   c_4 + k_{\mbox{\scriptsize nlp}}*(c_1+c_2) +
   k_{\mbox{\scriptsize wg}}*c_3.
\end{equation}
We then define the ordering as:
\begin{equation}
\langle c_1, c_2, c_3, c_4\rangle \prec \langle c'_1, c'_2,
c'_3, c'_4\rangle \mbox{ iff } \mbox{\it total\/}(\langle c_1, c_2, c_3, c_4\rangle) <
\mbox{\it total\/}(\langle c'_1, c'_2, c'_3, c'_4\rangle).
\end{equation}

\subsection{Comparison with Dijkstra's algorithm}
\label{robalt}
In a previous version of the implementation we used a generalised version
of {\sc Dijkstra}'s algorithm \cite{dijkstra}, \cite{nilsson71}, 
\cite{algorithms}, instead of the {\sc
  dag-shortest-path} presented above. Dijkstra's algorithm is more
general in that it is not restricted to acyclic graphs. On the other
hand, however, Dijkstra's algorithm requires that weights on edges are
positive (paths can only get worse if they are extended).  A potential
advantage of Dijkstra's algorithm for our purposes is that the
algorithm often does not have to investigate all edges. If edges
are relatively expensive to compute, then Dijkstra's algorithm might
turn out to be faster.

For instance, we can obtain a modest increase in efficiency by
exploiting Dijkstra's algorithm if we delay some of the work the
parser has to do for some category $q$, until the robustness component
actually has a need for that category $q$. Since Dijkstra's algorithm
will not visit every $q$, the amount of work is reduced.  We exploited
this in our implementation as follows. The parser works in two phases.
In the first phase (recognition) no semantic constraints are taken
into account (in order to pack all ambiguities). In the second phase
semantic constraints are applied. This second phase can then be
delayed for some category $q$ until Dijkstra's algorithm uses an edge
based on $q$. For a number of categories, therefore, this second phase
can be skipped completely.

However, we had three reasons for preferring the {\sc
  dag-shortest-path} algorithm given above. Firstly, this algorithm is
simpler than Dijkstra's algorithm. Secondly, negative weights do show
up in a number of circumstances. And thirdly, the expected efficiency
gain was not observed in practice.

An example where negative weights may show up is the following.
Suppose we define a method which takes into account Ngram scores but
nothing else, i.e. all other sources of information such as acoustic
scores are ignored. It turns out that a straightforward implementation
of such a method is non-optimal since it will favour paths in the
word-graph consisting of a small number of long words over paths (of
the same duration) consisting of a larger number of smaller words, 
only because more scores have to be added. A simple and effective
way to eliminate this effect, is to subtract a constant from
each score.  However, this subtraction may yield negative numbers.

\subsection{Best-first methods}
\label{bfm}
Rather than integrating parsing and disambiguation of the word-graph
as a single procedure, as we proposed above, it is also possible to
try to disambiguate the word-graph first, and then use the parser to
parse the resulting path in the word-graph.

We have implemented two versions of this approach. Both versions use
the search algorithm discussed above, by applying a method which takes
into account the acoustic scores and Ngram scores.  One version uses
$N=2$, the other version uses $N=3$. In section~\ref{eval} we refer to
these two methods as {\em best\_1\_bigram} and {\em best\_1\_trigram}
respectively. 

We have experimented with such methods in order to evaluate the
contribution of grammatical analysis to speech recognition. If, for
instance, the integrated method {\em nlp\_speech\_trigram} performs
significantly better than {\em best\_1\_trigram} then we can conclude
that grammatical analysis improves speech recognition. The results
below, however, do not permit this conclusion.

\section{Evaluation}
\label{eval}

We present a number of results to indicate how well the NLP component
currently performs. In the NWO Priority Programme, two alternative natural 
language processing modules are developed in parallel: the
`grammar-based' module described here, and a `data-oriented'
(statistical, probabilistic, DOP) module.  Both of these modules fit 
into the system architecture of OVIS. The DOP approach is documented
in a number of publications \cite{dopscha90,bonnema-acl97,tst38}. 

In order to compare both NLP modules, a formal evaluation has been
carried out on 1000 new, unseen, representative word graphs (obtained
using the latest version of the speech recognizer).  Full details on the evaluation procedure, and all evaluation results, are described elsewhere \cite{tst46,tst57}.
For these word
graphs, annotations were provided by our project partners consisting of
the actual sentences ('test sentences'), and updates ('test updates').

The Ngram models used by our implementation were constructed on the
basis of a corpus of almost 60K user utterances (almost 200K words). 

Some indication of the difficulty of the test-set of 1000 word-graphs
is presented in table~\ref{een}, both for the input word-graphs and
for the normalised word-graphs. The table lists the number of
transitions, the number of words of the actual
utterances, the average number of transitions per word, the average number
of words per graph, the average number of transitions per graph, and
finally the maximum number of transitions per graph. 
The number of transitions per word in the normalized word-graph is an
indication of the additional ambiguity that the parser encounters in
comparison with parsing of ordinary strings.

\begin{table}
\caption{\label{een}
{\bf Characterization of test set (1).}
}
\begin{center}
\begin{tabular}{crrrrrrr}\hline
& graphs & transitions & words & t/w & w/g &  t/g & max(t/g)\\\hline
input 
& 1000  &  48215       & 3229 & 14.9 & 3.2 & 48.2 & 793 \\
normalized
& 1000  &  73502       & 3229 & 22.8 & 3.2 & 73.5 &2943 \\     
\hline
\end{tabular}
\end{center}
\end{table}

A further indication of the difficulty of this set of word-graphs is
obtained if we look at the word and sentence accuracy obtained by a
number of simple methods. 
The string comparison on which sentence accuracy and word accuracy are
based is defined by the minimal number of substitutions, deletions and
insertions that is required to turn the first string into the second
(Levenshtein distance $d$).  Word accuracy is defined as
$ 1 -  \frac{d}{n} $ where $n$ is the length of the actual utterance.

The method {\em speech} only takes into account the acoustic scores
found in the word-graph. The method {\em possible}
assumes that there is an oracle which chooses a path such that it
turns out to be the best possible path.  This method can be seen as a
natural upper bound on what can be achieved.  The methods {\em bigram} 
({\em trigram}) report on a method which {\em only} uses a bigram (trigram) 
language model. The methods {\em speech\_bigram} ({\em
  speech\_trigram}) use a combination of bigram (trigram)
statistics and the speech score. 
\begin{table}
\caption{\label{two}{\bf Characterization of test set (2).}
}
\begin{center}
\begin{tabular}{crr}\hline
method               &        WA &      SA\\\hline
speech               &      69.8 &  56.0\\
possible             &      90.4 &  83.7\\
bigram               &      69.0 &  57.4\\
trigram              &      73.1 &  61.8\\
speech\_bigram       &      81.1 &  73.6\\
speech\_trigram      &      83.9 &  76.2\\\hline
\end{tabular}
\end{center}
\end{table}

\subsection{Efficiency}

Table~\ref{three} reports on the efficiency of the NLP components for
the set of 1000 wordgraphs and test utterances. The first two rows
present the results for sentences; the remaining rows provide the
results for word-graphs.  Listed are respectively the average number
of milliseconds per input; the maximum number of milliseconds; and the
maximum space requirements (per word-graph, in Kbytes). 

For most word-graphs we used the nlp\_speech\_trigram method as
described above. For large word-graphs (more than 100 transitions), we
first selected the best path in the word-graph based on acoustic
scores and N-gram scores only. The resulting path was then used as
input for the parser.  In the case of these large word-graphs, N=2
indicates that bigram scores were used, for N=3 trigram scores were
used. 

CPU-time includes tokenizing the word-graph, removal of pause
transitions, lexical lookup, parsing, the robustness/disambiguation
component, and the production of an update expression. \footnote{For
  the grammar-based methods, CPU-time was measured on a HP 9000/780
  machine running HP-UX 10.20, with SICStus Prolog 3 patch level 3.
  The statistics for the data-oriented module were obtained on a
  Silicon Graphics Indigo with a MIPS R10000 processor, running IRIX
  6.2.}

\begin{table}
\caption{\label{three}{\bf Efficiency (1).}}
\begin{center}
\begin{tabular}{llrrr}\hline
input         & method            & mean msec &  max msec & max Kbytes\\
\hline
test sentence & data-oriented     &  91       &      8632 &     14064 \\
test sentence & grammar-based     &  28       &       610 &       524 \\
\hline
word graphs   & data-oriented     &7011       &    648671 &    619504 \\
word graphs   & grammar-based N=2 & 298       &     15880 &      7143 \\
word graphs   & grammar-based N=3 &1614       &    690800 &     34341 \\
\hline
\end{tabular}
\end{center}
\end{table}

For word-graphs the average CPU-times are actually quite
misleading because CPU-times vary enormously for different
word-graphs.  For this reason, we present in table~\ref{four} the
proportion of word-graphs (in \%) that can be treated by the NLP
component within a given amount of CPU-time (in milliseconds).

\begin{table}
\caption{\label{four}{\bf Efficiency (2).}
}
\begin{center}
\begin{tabular}{lrrrrr}\hline
method            &  100 &   500 &  1000 & 5000  &10000 \\\hline
data-oriented     & 52.7 &  70.8 &  76.6 & 90.6  & 94.2 \\
grammar-based N=2 & 58.6 &  87.0 &  94.6 & 99.5  & 99.8 \\
grammar-based N=3 & 58.5 &  78.9 &  87.3 & 96.7  & 98.7 \\
\hline
\end{tabular}
\end{center}
\end{table}

\subsection{Accuracy}

The results for word accuracy given above provide a measure for the
extent to which linguistic processing contributes to speech
recognition. However, since the main task of the linguistic component
is to analyze utterances semantically, an equally important measure is
{\em concept accuracy}, i.e.\ the extent to which semantic analysis
corresponds with the meaning of the utterance that was actually
produced by the user.

For determining concept accuracy, we have used a semantically
annotated corpus of 10K user responses.  Each user response was
annotated with an {\em update} representing the meaning of the
utterance that was actually spoken.  The annotations were made by our
project partners in Amsterdam, in accordance with the existing
guidelines \cite{tst24}.

Updates take the form described in section~\ref{semantics}.  An update
is a logical formula which can be evaluated against an information
state and which gives rise to a new, updated information state. The
most straightforward method for evaluating concept accuracy in this
setting is to compare (the normal form of) the update produced by the
grammar with (the normal form of) the annotated update. A major
obstacle for this approach, however, is the fact that very
fine-grained semantic distinctions can be made in the update-language.
While these distinctions are relevant semantically (i.e. in certain
cases they may lead to slightly different updates of an information
state), they can often be ignored by a dialogue manager. For instance,
the two updates below are semantically not equivalent, as the ground-focus distinction is slightly
different.
\begin{numbered}\begin{verbatim}
user.wants.destination.place.([# town.leiden];[! town.abcoude])
user.wants.destination.([# place.town.leiden];[! place.town.abcoude])
\end{verbatim}\end{numbered}
\noindent However, the dialogue manager will decide in both cases that
this is a correction of the destination town. 

Since semantic analysis is the input for the dialogue manager, we have
measured concept accuracy in terms of a simplified version
of the update language. Inspired by a similar proposal in Boros et
al.\/ \shortcite{Boros}, we
translate each update into a set of {\em semantic units}, where a unit
in our case is a triple $\langle${\tt CommunicativeFunction, Slot,
  Value}$\rangle$. For instance, the two examples above both translate as 

\noindent
$\langle$ {\tt denial, destination\_town, leiden} $\rangle$\\
$\langle$ {\tt correction, destination\_town, abcoude} $\rangle$

\noindent
Both the updates in the annotated corpus and the updates produced by
the system were translated into semantic units.

Semantic accuracy is given in table~\ref{five} according to four
different definitions. Firstly, we list the proportion of utterances
for which the corresponding semantic units exactly match the semantic
units of the annotation ({\em match}). Furthermore we calculate 
{\em precision} (the number of correct semantic units divided by the
number of semantic units which were produced) and {\em recall} 
(the number of correct semantic units divided by the number of semantic
units of the annotation). 
Finally, following Boros et al.\/ \shortcite{Boros}, we also present concept accuracy as

\[
CA = 100 \left( 1 - \frac{SU_S + SU_I + SU_D}{SU} \right) \%
\] 

\noindent 
where $SU$ is the total number of semantic units in the translated corpus
annotation, and $SU_S$, $SU_I$, and $SU_D$ are the number of
substitutions, insertions, and deletions that are necessary to make
the translated grammar update equivalent to the translation of the 
corpus update. 

We achieve the results listed in table~\ref{five} for the test-set of
1000 word-graphs.  String accuracy is presented in terms of
word-accuracy (WA) and sentence accuracy (SA).  

\begin{table}
\caption{\label{five}{\bf Accuracy.}}
\begin{center}
\begin{tabular}{ccrrrrrr}\hline
Input         &     Method & \multicolumn{2}{c}{String accuracy} 
                           & \multicolumn{4}{c}{Semantic accuracy}\\  
              &                   &  WA   & SA  & match & precision & recall &   CA\\\hline
test sentence & data-oriented     & N/A  & N/A  & 93.0  & 94.0      & 92.5   & 91.6\\
test sentence & grammar-based     & N/A  & N/A  & 95.7  & 95.7      & 96.4   & 95.0\\
\hline
word-graph    & data-oriented     & 76.8 & 69.3 & 74.9  & 80.1      & 78.8   & 75.5\\
word-graph    & grammar-based N=2 & 82.3 & 75.8 & 80.9  & 83.6      & 84.8   & 80.9\\
word-graph    & grammar-based N=3 & 84.2 & 76.6 & 82.0  & 85.0      & 86.0   & 82.6\\
\hline
\end{tabular}
\end{center}
\end{table}

\section*{Conclusion}

The results given above lead to the following conclusions.

\begin{itemize}
\item Sophisticated grammatical analysis is fast enough for practical
  spoken dialogue systems.
\item Moreover, grammatical analysis is effective for the purposes of
  the present application. Almost all user utterances can be analysed
  correctly. This is somewhat surprising. Typically, linguistic
  ambiguities are a major obstacle for practical NLP systems. The
  current application is very simple in the sense that such linguistic
  ambiguities do not seem to play a significant role. The
  ambiguities introduced by the speech recognizer (as multiple paths
  in the word-graph) are a far more important problem.
\item Grammatical analysis does not seem to help much to solve
  the problem of disambiguating the word-graph. The best method
  incorporating grammatical analysis performs about as well as a
  method which solely uses N-gram statistics and acoustic scores for
  disambiguation of the word-graph. However, in the latter case
  grammatical analysis of the type proposed here is useful in
  providing a robust analysis of the best path.
\end{itemize}

We have argued in this paper that sophisticated grammatical analysis
in combination with a robust parser can be applied successfully as an
ingredient of a spoken dialogue system.  Grammatical analysis
is thereby shown to be a viable alternative to techniques such as concept
spotting. We showed that for a state-of-the-art application (public
transport information system) grammatical analysis can be applied
efficiently and effectively. It is expected that the use of
sophisticated grammatical analysis will allow for easier construction of
linguistically more complex spoken dialogue systems.

\section*{Acknowledgements}
This research was carried out within the framework of the Priority
Programme Language and Speech Technology (TST). The TST-Programme is
sponsored by NWO (Dutch Organization for Scientific Research).

\bibliography{refs}

\label{lastpage}

\end{document}